\pdfoutput=1
\documentclass[10pt,twocolumn,letterpaper]{article}

\usepackage[pagenumbers]{cvpr}

\usepackage{multirow}
\usepackage{colortbl}
\usepackage{enumitem}
\usepackage{microtype}

\DeclareMathOperator*{\argmin}{arg\,min}
\newcommand{\method}{MOSAIC\xspace}
\newcommand{\best}[1]{\textbf{#1}}

\definecolor{cvprblue}{rgb}{0.21,0.49,0.74}
\usepackage[breaklinks,colorlinks,allcolors=cvprblue,
  pdfauthor={Zeyu Chen, Enkeleda Thaqi, Zhuojiang Cai, Enkelejda Kasneci},
  pdftitle={Generative Video Anonymization: Whole-Body Identity Suppression with Scene and Action Preservation}]{hyperref}

\title{Generative Video Anonymization:\\Whole-Body Identity Suppression with Scene and Action Preservation}

\author{Zeyu Chen \quad Enkeleda Thaqi \quad Zhuojiang Cai \quad Enkelejda Kasneci\\
Technical University of Munich\\
Munich Center for Machine Learning (MCML)}

\begin{document}
\maketitle
\begin{abstract}
Re-identification systems recognize people by body shape, clothing, hair and gait, so anonymizing only the face leaves most of a person's identity in a video. We study \emph{Generative Video Anonymization} (GVA): replacing every person in a video with a different person who performs the same action in the same scene. We formalize GVA as identity suppression under action, scene and temporal constraints, and argue that general video editing methods are ill-suited to it because they do not ground the subject, do not push the identity away, and do not anchor the new identity over time. We propose an evaluation protocol that attacks anonymized videos with temporally disjoint retrieval and with appearance, clothes-changing, face, gait and soft-biometric attackers, and that measures utility on time-aligned frames. We introduce Belvedere Museum, an egocentric benchmark recorded during regular museum visits, where visitors must be distinguished from painted and sculpted people. Finally, we present \method, a training-free pipeline that grounds people with LMM-verified segmentation, plans a counterfactual identity per clip, and regenerates the people with a mask-conditioned video diffusion transformer. On OpenHumanVid, where the original video is re-identified at 95\% cross-time Rank-1, \method reduces appearance, clothes-changing, face and gait attackers to 24.5, 19.5, 4.9 and 42.6, the lowest of any temporally coherent generative method, while preserving the scene at 0.951 SPS.
\end{abstract}

\section{Introduction}
\label{sec:intro}

Video of people is central to behaviour analysis, egocentric perception and human-centric generation, yet releasing it is constrained by privacy. Identity in video is not confined to the face: re-identification (Re-ID) systems match people across cameras by body shape, clothing, hair and gait~\cite{he2021transreid,ye2021deep,chao2019gaitset,gu2022clothes}, and they keep working when the face is hidden~\cite{oh2016faceless}. Face anonymization therefore leaves most of the identity signal in place.

Existing alternatives trade one problem for another. Blurring or masking the whole person removes identity, but also the action and the scene interaction that make the video useful~\cite{ryoo2017privacy}. Generative face anonymization~\cite{maximov2020ciagan,hukkelas2019deepprivacy,hukkelas2023deepprivacy2} is realistic but changes only the face. What is needed is to separate the identity of a person from what the person does: the anonymized video should show a \emph{different} person performing the \emph{same} action in the \emph{same} scene.

We call this task \textbf{Generative Video Anonymization (GVA)}. It is a generative problem with an unusual objective: the person must move far from the source in a biometric feature space while the action, the scene and the temporal coherence stay fixed, so pixel fidelity to the source is the wrong target inside the person region (\cref{sec:problem}). General video editors~\cite{wu2023tune,geyer2024tokenflow,wei2026univideo} can change appearance on instruction, but they do not know which pixels carry identity, they optimize agreement with the instruction rather than identity distance, and they do not anchor the new identity over time.

\begin{figure}[t]
  \centering
  \includegraphics[width=\linewidth]{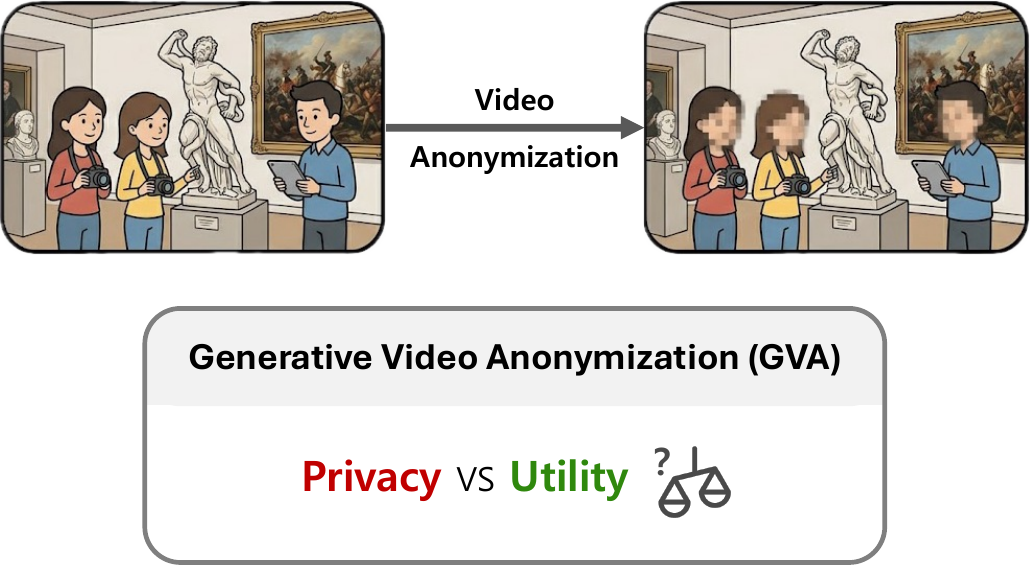}
  \caption{\textbf{Generative Video Anonymization.} Each person is replaced by a different person who keeps the same action and scene, so privacy and utility have to be balanced.}
  \label{fig:teaser}
\end{figure}

\begin{figure}[t]
  \centering
  \includegraphics[width=0.75\linewidth]{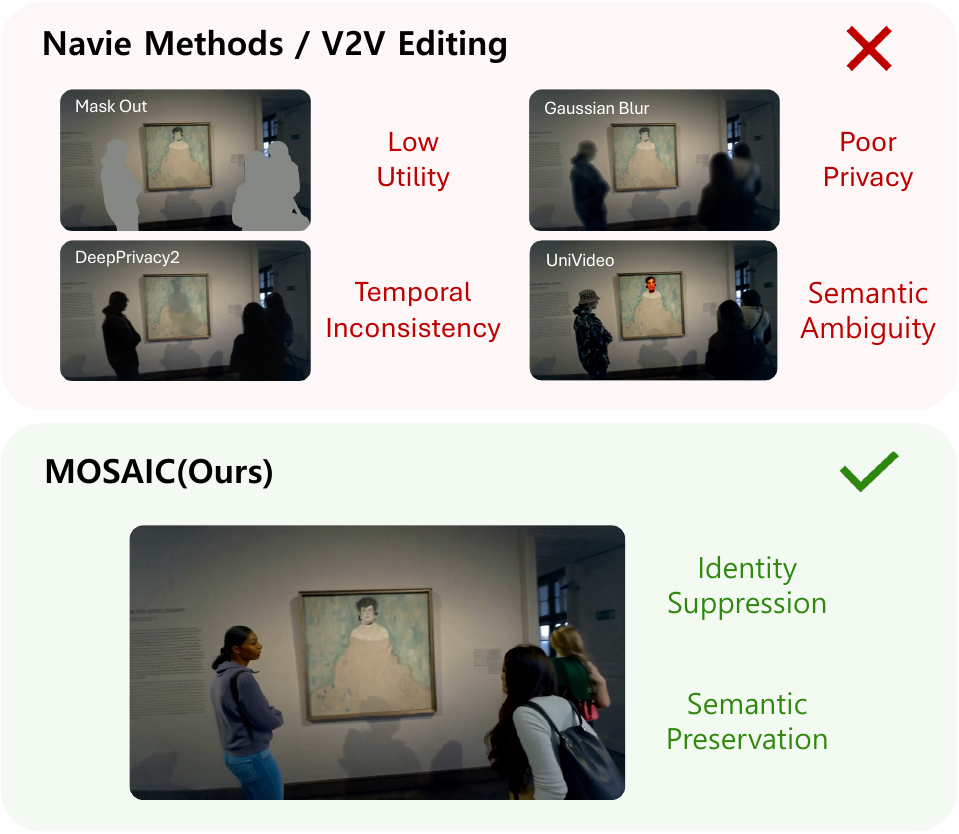}
  \caption{\textbf{Why existing edits fall short.} Masking and blurring remove the action. DeepPrivacy2 (full body) is temporally inconsistent. General video editing neither localizes the person precisely nor avoids semantic errors. \method suppresses identity and leaves the scene in place.}
  \label{fig:teaser_compare}
\end{figure}

\noindent Our contributions are:
\begin{itemize}[leftmargin=*,itemsep=2pt,topsep=2pt]
  \item \textbf{Task.} We formalize GVA as identity suppression under action, scene and temporal constraints and identify three gaps that make general video editing ill-suited to it (\cref{sec:problem}).
  \item \textbf{Protocol.} We evaluate privacy with temporally disjoint retrieval and attackers that read different cues (appearance and clothes-changing Re-ID, face, gait, soft biometrics), and utility on time-aligned frames, with scene preservation outside the person, the warping error in excess of the source, action and pose agreement, and realism (\cref{sec:eval_protocol}).
  \item \textbf{Benchmark.} We introduce Belvedere Museum, egocentric video recorded with eye-tracking glasses during regular museum visits, where visitors must be told apart from painted and sculpted people (\cref{sec:datasets}).
  \item \textbf{Method.} We present \method, a training-free pipeline that grounds every present person with LMM verification, plans a counterfactual identity, and regenerates the people with a mask-conditioned video diffusion transformer (\cref{sec:method}).
\end{itemize}
On OpenHumanVid, where the original video is re-identified at 95\% cross-time Rank-1, \method reduces appearance, clothes-changing, face and gait attackers to 24.5, 19.5, 4.9 and 42.6 while preserving the scene at 0.951 SPS, the strongest balance of privacy and utility among temporally coherent generative methods.

\section{Related work}
\label{sec:related}

\paragraph{Face and body anonymization.}
Obfuscation by blurring or masking removes identity together with utility~\cite{ryoo2017privacy,mcpherson2016defeating}. Generative face anonymization replaces faces with synthetic ones while keeping pose and expression, with GANs~\cite{hukkelas2019deepprivacy,maximov2020ciagan,sun2018natural} and diffusion models~\cite{kim2023dc,liu2023diffprotect}. DeepPrivacy2~\cite{hukkelas2023deepprivacy2} extends this to the full body, but treats every frame independently. The concurrent ReGenHuman~\cite{sun2026regenhuman} uses the same inpainting backbone as \method, but its masks are not restricted to physically present people, and it has no counterfactual identity objective. The difference is Stage~I and Stage~II of \method. Most of these methods are evaluated by face recognition only, whereas Re-ID systems match people by body shape, clothing and gait~\cite{ye2021deep,wang2024survey,chao2019gaitset}. We therefore evaluate against appearance, clothes-changing, face, gait and attribute attackers.

\paragraph{Video editing and inpainting.}
Text-driven video editors adapt image diffusion models to video~\cite{wu2023tune,geyer2024tokenflow,zhang2024controlvideo,khachatryan2023text2video}, and unified models such as UniVideo~\cite{wei2026univideo} and VACE~\cite{jiang2025vace} support instruction editing and mask-conditioned inpainting. Character animation synthesizes a controllable person from a reference image~\cite{hu2024animate,wang2024mimo}, which is the opposite objective of anonymization. None of these methods decides which pixels carry identity or what the replacement identity should be. \method supplies both and uses VACE only as the inpainting backbone.

\paragraph{Grounding and evaluation.}
Promptable segmentation~\cite{kirillov2023segment,carion2026sam} and video trackers~\cite{cheng2022xmem,yang2023track} mask anything person-like, including painted and sculpted people. We verify each candidate with an LMM~\cite{qwen3vl} so that only physically present people are anonymized. Prior evaluations typically report one Re-ID or face matcher, often with query and gallery from the same moment. We add temporally disjoint retrieval, attackers that read different cues, and utility measures on motion and pose rather than pixels.

\section{Problem formulation}
\label{sec:problem}

Let $V = \{I_t\}_{t=1}^{T}$ be a video containing identifiable people. Generative Video Anonymization (GVA) seeks a transformation $\mathcal{T}$ whose output $\hat V = \mathcal{T}(V)$ (i) suppresses whole-body biometric identity, (ii) preserves behaviour and scene, and (iii) remains temporally realistic:
\begin{equation}
\begin{aligned}
  \mathcal{T}^{*} = \argmin_{\mathcal{T}}\; & \mathcal{L}_\mathrm{id}(V, \hat V) \\
  \text{s.t.}\;\; & \mathcal{C}_\mathrm{act}(\hat V, V) \ge \gamma,\;\; \mathcal{C}_\mathrm{bg}(\hat V, V) \ge \eta,\\
  & \mathcal{C}_\mathrm{temp}(\hat V) \ge \tau .
\end{aligned}
\label{eq:gva}
\end{equation}
$\mathcal{L}_\mathrm{id}$ is the identity leakage measured by an attacker in a biometric feature space, and $\mathcal{C}_\mathrm{act}$, $\mathcal{C}_\mathrm{bg}$, $\mathcal{C}_\mathrm{temp}$ measure the preservation of action, background and temporal coherence, with acceptance thresholds $\gamma, \eta, \tau$. \Cref{sec:eval_protocol} instantiates each term. We leave the thresholds free, since the acceptable operating point depends on the release scenario, and report the full privacy--utility trade-off (\cref{fig:tradeoff}) instead of a single feasible point.

The objective and the constraints live in different spaces. Suppressing identity requires the person to move away from the source in the attacker's feature space, whereas utility is defined by motion, scene and stability. Fidelity measures computed against $V$ inside the person region therefore cannot serve as utility measures for GVA: they penalize exactly the change that anonymization requires. We measure scene fidelity only outside the person masks and compare motion and actions rather than pixels.

\subsection{Why video editing methods fall short}
\label{sec:v2v_analysis}

General video-to-video (V2V) editors such as UniVideo~\cite{wei2026univideo} can change a person's appearance from a text instruction, but three properties of their objective conflict with \cref{eq:gva}.

\paragraph{F1: Spatial grounding.} Instructions act on the whole frame. Without a subject-level mask, edits spread into the background (violating $\mathcal{C}_\mathrm{bg}$), and people the instruction does not single out, such as bystanders, may stay untouched.

\paragraph{F2: Identity objective.} Editors optimize agreement with the instruction, not distance in a Re-ID feature space. A visible change, \eg a new jacket colour, can leave body shape, hair, gait and the face intact, so $\mathcal{L}_\mathrm{id}$ stays high.

\paragraph{F3: Temporal identity.} The new appearance is not anchored across frames, so a synthesized person may drift or flicker over the clip, violating $\mathcal{C}_\mathrm{temp}$.

\method (\cref{sec:method}) addresses each gap with one stage, and \cref{sec:experiments} measures each gap separately.

\begin{figure*}[t]
  \centering
  \includegraphics[width=0.97\textwidth]{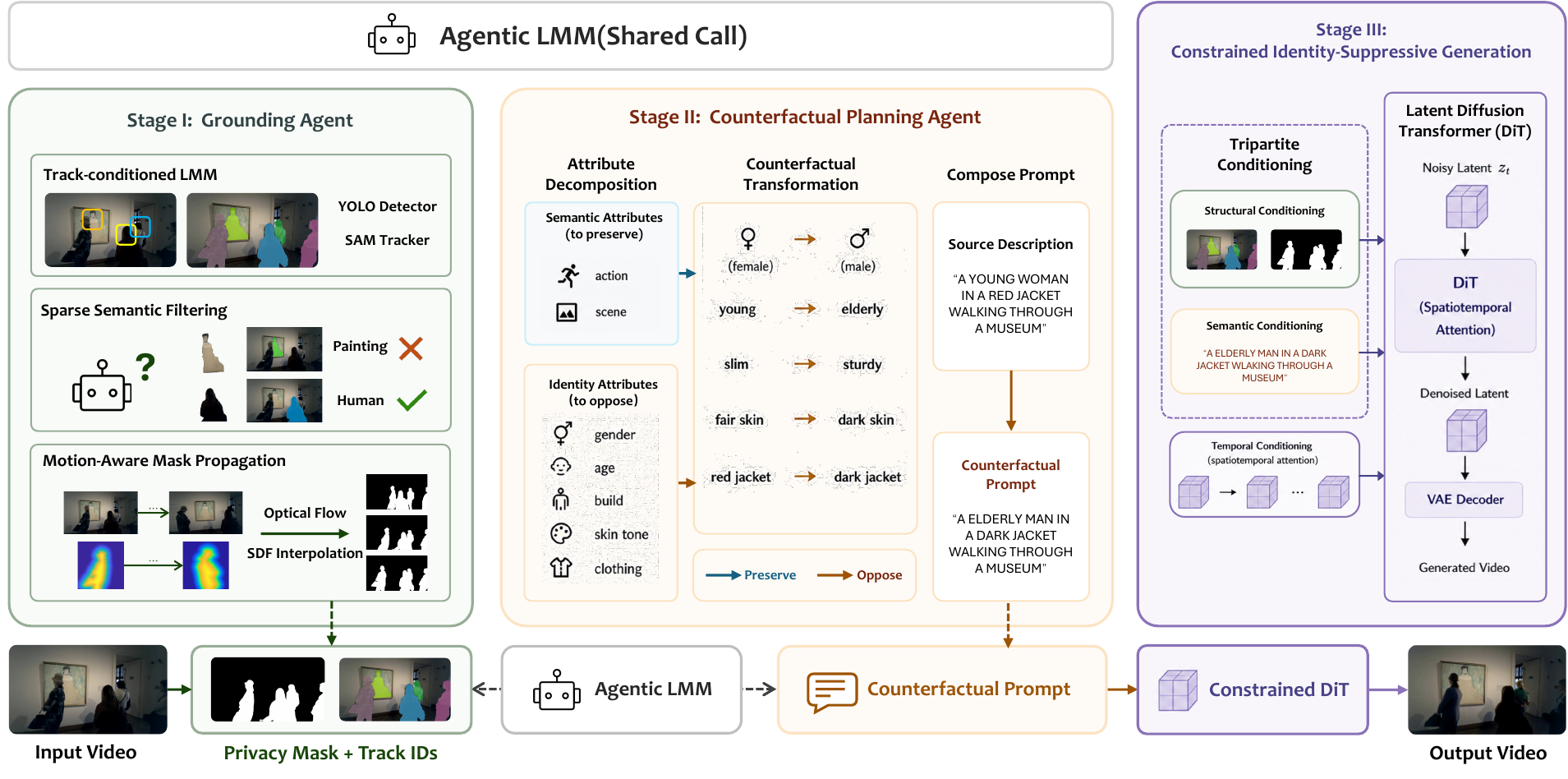}
  \caption{\textbf{Overview of \method.} Stage~I grounds every person in the clip (SAM3 candidates verified by an LMM, then temporally completed). Stage~II plans a counterfactual identity for the clip, and Stage~III regenerates the masked people with a mask-conditioned video diffusion transformer. All components are pretrained and used without fine-tuning.}
  \label{fig:mosaic_overview}
\end{figure*}

\section{\method}
\label{sec:method}

\method (Mask-Oriented Synthesis via Identity Counterfactuals) addresses the three gaps of \cref{sec:v2v_analysis} with one stage each (\cref{fig:mosaic_overview}). It is training-free: every stage uses an off-the-shelf model with fixed weights, and all settings below are shared by both datasets.

\subsection{Stage I: Grounding}
\label{sec:grounding}

Stage~I addresses F1. It must mark every physically present person, including bystanders, while leaving depicted people (paintings, sculptures, prints) and shadows untouched.

\paragraph{Candidates.}
We run the SAM3 video predictor~\cite{carion2026sam} with the text prompt ``humans'', which returns per-frame instance masks $\{\mathbf{m}^{(i)}_t\}$ with object identities $i$ that persist across frames. SAM3 responds to anything that looks like a person, so these candidates include artwork and shadows.

\paragraph{LMM verification.}
A large multimodal model (LMM) verifies each candidate in each frame. Shown the frame with the candidate highlighted and a crop around it, the model classifies the candidate as (A) a physically present person, (B) a person depicted in a flat artwork, (C) a cast shadow, or (D) uncertain. Restricting the next-token distribution to these four answers gives a distribution over verdicts,
\begin{equation}
  p^{(i)}_t(c) = \operatorname{softmax}_{c \in \{A,B,C,D\}} \ell^{(i)}_t(c),
\end{equation}
and we keep a candidate if $\arg\max_c p^{(i)}_t(c) \in \{A, D\}$. Keeping uncertain candidates is deliberate: a missed person leaks an identity, whereas an extra mask only costs some background.

\paragraph{Temporal completion.}
Tracks drop out under the motion blur and occlusion typical of egocentric video. When instance $i$ is missing between two observations $t_a < t_b$ with $t_b - t_a - 1 \le 20$, we interpolate it: both endpoint masks are translated to the linearly interpolated centre $\mathbf{c}_t = (1-\alpha)\mathbf{c}_{t_a} + \alpha\mathbf{c}_{t_b}$, $\alpha = \frac{t - t_a}{t_b - t_a}$, and the missing mask is the positive set of the blended signed distance fields~\cite{osher1988fronts} of the translated masks,
\begin{equation}
  \phi_t = (1-\alpha)\,\phi_{t_a}(\mathbf{x} + \mathbf{c}_{t_a} - \mathbf{c}_t) + \alpha\,\phi_{t_b}(\mathbf{x} + \mathbf{c}_{t_b} - \mathbf{c}_t).
\end{equation}
The interpolated mask is refined with optical flow~\cite{farneback2003two} from the nearer endpoint and cleaned morphologically, and every mask is slightly dilated, giving the per-frame privacy mask $M_t$.

\subsection{Stage II: Counterfactual identity planning}
\label{sec:planning}

Stage~II addresses F2. A generic caption tells the generator what the person does, but not who they should become, so the generator tends to reproduce the source appearance from the visible context. Stage~II instead specifies an identity that differs from the source in every attribute a Re-ID model relies on.

\paragraph{Attribute extraction.}
Once per clip, an LMM describes four evenly spaced keyframes, merges the four descriptions into one, and extracts the identity attributes $\mathbf{a}_\mathrm{id} = \{$gender, age group, body type, skin tone, clothing$\}$. A smaller LMM suffices for captioning, while the per-instance verification of Stage~I uses the larger one, since its errors cost privacy or artwork.

\paragraph{Counterfactual identity.}
A fixed mapping $\mathrm{CF}$ turns $\mathbf{a}_\mathrm{id}$ into a target identity $\tilde{\mathbf{a}}_\mathrm{id}$: the target gender is the other binary label, and every other attribute is resampled away from the source (a different age bracket, the opposite body-type group, another skin tone, another outfit). The draws are seeded by the clip, so each clip keeps a fixed target identity while targets differ across clips. Sampling, rather than a deterministic opposite, keeps the synthetic population diverse instead of mapping every source to one stereotype. The target gender is only a prompt constraint: the mask silhouette carries body shape, so the rendered person often keeps the source presentation, and privacy is measured by retrieval, not by a visible sex change.

\paragraph{Prompt.}
The generation prompt $P$ consists of a sentence stating $\tilde{\mathbf{a}}_\mathrm{id}$, a fixed instruction to keep the action, pose, motion, viewpoint, timing, lighting and every non-person region, and a generic scene caption produced by an LMM. The extracted source attributes are not part of $P$, although the unconstrained caption occasionally echoes one (\cref{sec:limitations}).

\subsection{Stage III: Mask-constrained generation}
\label{sec:generation}

Stage~III addresses F3 by regenerating all masked people jointly over the whole clip with a video diffusion transformer, so the new identity is synthesized once for the clip rather than frame by frame. We use Wan2.1-VACE-1.3B~\cite{wan2025wan,jiang2025vace} in its inpainting mode. Its context input is the source video with every masked pixel replaced by mid-gray, $V \odot (1-M) + 0.5\,M$, together with the mask video $M$, and the prompt $P$ enters through cross-attention. Because the people are removed from the context, the model cannot copy their appearance and must synthesize them from $P$, constrained by the surrounding scene and the mask silhouettes, which carry the body pose and motion.

We generate up to 81 frames at 480p ($832{\times}480$ for 16:9 clips). Clips longer than 81 frames are resampled to 81 frames spanning their full duration, and shorter clips keep their frames. No parameter is fine-tuned.

\section{Evaluation protocol}
\label{sec:eval_protocol}

A GVA method must defeat identity attackers while keeping the video useful, so the protocol pairs a set of attackers with a set of utility measures. Both are computed on fixed cohorts and on time-aligned frames, with 95\% bootstrap confidence intervals over clips (10{,}000 resamples~\cite{efron1993bootstrap}).

\paragraph{Cohorts and alignment.}
Belvedere uses its 49 test clips. For OpenHumanVid (OHV) we draw 50 clips (seed 42) from the 256 test clips for which every baseline produced an output, and freeze the list. Every method is evaluated on the same clips, and a missing output reduces that method's coverage and is never replaced by another clip.

Generative methods change the length, frame rate and framing of a clip, so frame $j$ of an output generally does not show frame $j$ of the source. \method spreads at most 81 frames over the whole clip and center-crops to $832{\times}480$. ReGenHuman~\cite{sun2026regenhuman} keeps every third frame (at most 49) and stretches the full frame. UniVideo~\cite{wei2026univideo} conditions only on the first frames of long clips and crops to its working resolution. For every output frame we therefore recover the source frame and the source region it shows. Both follow from each method's code, except UniVideo's timing, which we recover by dynamic time warping on the unmasked background. All metrics use the \emph{common span} $[0,S]$ of source frames that every method covers for that clip, and the source is cropped to the region visible in each output before comparison, so no method is compared on content it was never given. Attackers run on each video at its own resolution, and utility metrics compare frames resized to $832{\times}480$. The frame maps and crop boxes are released with the code.

\subsection{Identity attackers}

\paragraph{Retrieval setting.}
Each clip is one identity. Persons are detected by YOLOv8n~\cite{jocher2023yolov8} (confidence $\ge 0.5$) in 20 source frames $s_k = \operatorname{round}(kS/19)$, $k=0,\dots,19$, and in the output frames that show them. Queries are crops of the original clip, the gallery holds crops of the anonymized clips, and an attacker succeeds when a query retrieves its own clip. We report Rank-1 and mAP averaged over query clips, the mean cosine similarity $\mathcal{S}_{\cos}$ between original and anonymized crops of the same clip, the number of gallery identities $N$ and the Rank-1 chance level $1/N$. Persons that become undetectable leave the gallery, which counts as a miss for their query.

The usual protocol (\emph{same-time}) queries and searches all 20 frames, so a query can be matched against the anonymized crop of the very same moment, where pose, viewpoint and background agree. This measures how much of the \emph{instantaneous} appearance survives. Our main protocol is \emph{cross-time}: queries come from frames $k \in [0,7]$ (the first 40\% of the clip) and the gallery from frames $k \in [12,19]$ (the last 40\%), so the attacker must recognize the person at a different moment, as in real re-identification.

\paragraph{Attackers.}
We use attackers that read different identity cues:
\begin{itemize}[leftmargin=*,itemsep=1pt,topsep=2pt]
  \item \textbf{Appearance Re-ID.} OSNet~\cite{zhou2019omni} and OSNet-AIN~\cite{zhou2021learning} trained on MSMT17~\cite{wei2018msmt}, and TransReID~\cite{he2021transreid} trained on Market-1501~\cite{zheng2015scalable}. For comparison with earlier GVA evaluations we also report OSNet with ImageNet weights.
  \item \textbf{Clothes-changing Re-ID.} CSCI~\cite{pathak2025csci} with an EVA-02-L backbone trained on PRCC~\cite{yang2019prcc}, which is trained to ignore clothing and therefore relies on body shape, hair and face.
  \item \textbf{Face.} ArcFace~\cite{deng2019arcface} (InsightFace \texttt{buffalo\_l}, R50 trained on WebFace600K, SCRFD detector~\cite{guo2022scrfd}, detection score $\ge 0.5$) on the 20 full frames. We also report the face detection rate.
  \item \textbf{Gait.} GaitBase~\cite{fan2023opengait} trained on CASIA-B~\cite{yu2006casiab}, on YOLOv8n-seg silhouettes of the largest person in 40 frames (at least 8 valid silhouettes). Cross-time uses the first versus the second half.
  \item \textbf{Soft biometrics.} Qwen2-VL-2B~\cite{wang2024qwen2} labels gender, age group, upper and lower clothing colour and hair of the largest original person in four frames, and of the anonymized person at the same place (the matched detection, or the original box when nothing is detected, so that making a person undetectable does not hide it from this attacker). Per clip we take the majority label and report how often the anonymized label equals the original one, against the agreement expected from the two label marginals alone.
\end{itemize}

\subsection{Utility}

\paragraph{Scene preservation (SPS).}
With the Stage~I person mask $M_k$ of source frame $s_k$ and the matching output frame $\hat I_k$,
\begin{equation}
  \mathrm{SPS} = \frac{1}{20}\sum_{k} \mathrm{SSIM}\big((1-M_k)\odot I_k,\; (1-M_k)\odot \hat I_k\big),
\end{equation}
using SSIM~\cite{wang2004ssim}. SPS penalizes changes to the scene and uses the same mask for every method.

\paragraph{Temporal stability.}
On up to 21 consecutive output frames around the middle of the common span, at the method's own frame rate, we estimate RAFT flow~\cite{teed2020raft} and report the photometric warping error~\cite{lai2018learning}
\begin{equation}
  E_\mathrm{warp}(\hat V) = \frac{1}{T-1}\sum_{t} \frac{\lVert O_t \odot (\hat I_t \!-\! \mathcal{W}(\hat I_{t+1}, F_{t\to t+1}))\rVert_1}{3\,\lVert O_t\rVert_1},
\end{equation}
where $O_t$ masks pixels that fail a forward-backward consistency check~\cite{sundaram2010dense}. Natural videos have non-zero $E_\mathrm{warp}$ (lighting, blur, flow errors), so we report the \emph{excess} over the source, $\Delta E_\mathrm{warp} = E_\mathrm{warp}(\hat V) - E_\mathrm{warp}(V)$, with flow estimated on each video itself and $E_\mathrm{warp}(V)$ computed on the (cropped) source frames that the output frames show, so that both terms see the same frame spacing. A common shortcut, the mean squared difference between the flows of the source and the anonymized video, mixes motion change with flicker. The scale of $\Delta E_\mathrm{warp}$ depends on the source's own warping error, which differs by up to two orders of magnitude between the datasets, so values are comparable only within one dataset.

\paragraph{Action and pose.}
$\mathcal{A}_\mathrm{cons}$ is the cosine similarity of R(2+1)D-18~\cite{tran2018closer} Kinetics-400~\cite{kay2017kinetics} logits of 16 uniformly sampled frames, and we also report whether the top-1 action label is unchanged. Body pose is compared with YOLOv8n-pose~\cite{jocher2023yolov8}: persons detected in the original (confidence $\ge 0.5$) are matched to anonymized detections at IoU $\ge 0.5$, and we report the detection retention and the COCO OKS~\cite{lin2014coco} of matched keypoints.

\paragraph{Realism.}
Frame FID~\cite{heusel2017gans} between original and anonymized frames (20 per clip, pooled per dataset), and FVD computed on I3D Kinetics-400 logits after resampling every video to the output frames showing 81 evenly spaced source frames of the common span, at $832{\times}480$.

\section{Experiments}
\label{sec:experiments}

\begin{figure*}[t]
  \centering
  \includegraphics[width=\textwidth]{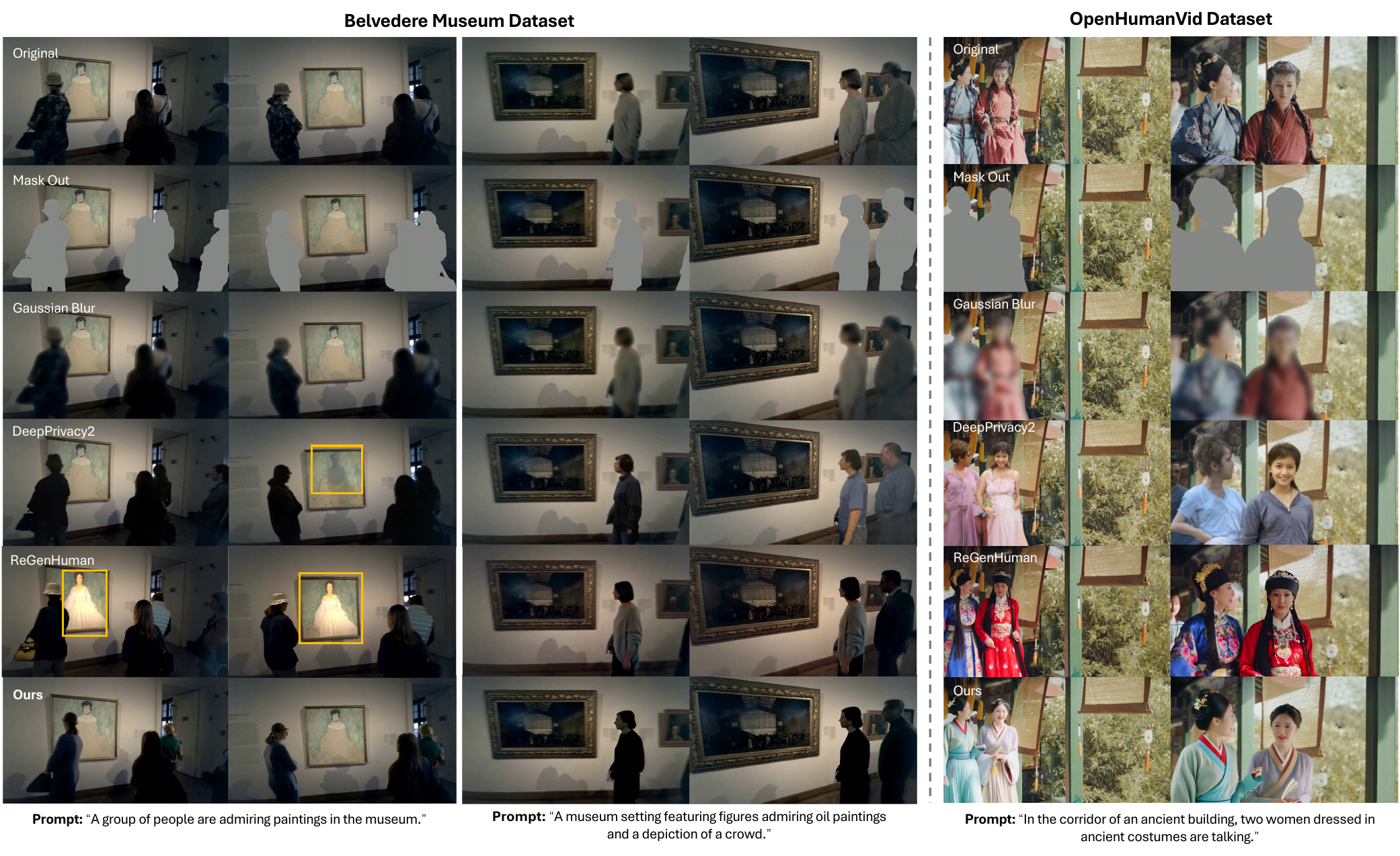}
  \caption{\textbf{Qualitative results.} Two frames from each of two Belvedere clips and one OpenHumanVid clip. Rows are the source, mask-out, Gaussian blur, DeepPrivacy2 (full body), ReGenHuman and \method. Yellow boxes mark a painted person that the method has altered. The sentence under each clip is the printed scene description.}
  \label{fig:qualitative}
\end{figure*}

\subsection{Setup}
\label{sec:datasets}

\paragraph{Datasets.}
\emph{OpenHumanVid (OHV)}~\cite{li2025openhumanvid} is a large human-centric video dataset of third-person footage. \method anonymizes its 256-clip test split, and we evaluate on the fixed cohort of \cref{sec:eval_protocol}.
\emph{Belvedere Museum} (ours) was recorded with the scene camera of Tobii Pro Glasses~\cite{TobiiProLab} during regular visits to the Belvedere Museum, Vienna. It contains two continuous egocentric recordings split into 618 segments, all anonymized by \method. The test clips span full-body, upper-body and close-up views and visitors of different ages and genders, and they include the hard cases of egocentric video: fast head motion, strong illumination contrast, and paintings and sculptures of people next to real visitors. We plan to release the clips with the Stage~I masks. See Ethics for the consent and release conditions.

\paragraph{Baselines.}
\emph{Blur} (Gaussian, kernel 99) and \emph{Mask-out} (gray fill) are applied inside the Stage~I mask, and they bound the trade-off without producing a person. \emph{DeepPrivacy2}~\cite{hukkelas2023deepprivacy2} is run in its full-body mode and, as a face-only reference, in its face mode. \emph{UniVideo}~\cite{wei2026univideo} (Qwen2.5-VL-7B with HunyuanVideo) receives the editing instruction produced by our Stage~II for the same clip, so it is asked for the same counterfactual identity. \emph{ReGenHuman}~\cite{sun2026regenhuman} regenerates the person with Wan2.1-VACE-1.3B and a released LoRA, from SAM3 person masks. It aborts on frames with an empty person mask, so its Belvedere numbers cover 37 of the 49 clips. Its masks are not restricted to physically present people and cover a painted sitter in \cref{fig:qualitative}.

\paragraph{Implementation.}
\method runs without training (\cref{sec:method}), and its LMMs are from the Qwen family~\cite{bai2023qwen,wang2024qwen2,qwen3vl}. Generation uses 50 UniPC steps, flow shift 16, classifier-free guidance 5, context scale 1 and seed 2025. All metrics use the same masks, frames and attackers for every method. An earlier variant, \emph{\method, free-text prompt}, replaces the structured prompt with Stage~II's free-form description. The structured prompt leaks less on OpenHumanVid (OSNet 24.5 versus 27.7), so it is the reference.

\subsection{Privacy and utility}

\begin{table*}[t]
\centering
\footnotesize
\setlength{\tabcolsep}{3pt}
\caption{\textbf{Identity leakage under cross-time retrieval} (Rank-1, \%, lower is better; queries from the first 40\% of each clip, gallery from the last 40\%). Attr.: soft-biometric labels unchanged by anonymization (\%), against chance 48.2 (Belvedere) and 44.6 (OHV). Chance Rank-1 is $1/N$ with $N\le 49$ and $N\le 50$ gallery identities; per-method $N$ and mAP are in the appendix. Bold: best among person-replacing methods.}
\label{tab:privacy}
\begin{tabular*}{\textwidth}{@{\extracolsep{\fill}}lcccccccccccc}
\toprule
 & \multicolumn{6}{c}{Belvedere} & \multicolumn{6}{c}{OpenHumanVid} \\ \cmidrule(lr){2-7} \cmidrule(lr){8-13}
Method & OSNet & TransReID & CSCI & ArcFace & Gait & Attr. & OSNet & TransReID & CSCI & ArcFace & Gait & Attr. \\
\midrule
Original & 29.4 & 32.4 & 33.1 & 26.2 & 18.8 & -- & 95.0 & 96.0 & 96.2 & 98.4 & 83.0 & -- \\
\midrule
\multicolumn{13}{@{}l}{\emph{Heuristic}} \\
Blur & 23.7 & 26.5 & 27.6 & 16.7 & 22.9 & 31.9 & 78.7 & 80.4 & 85.1 & 68.2 & 61.7 & 48.8 \\
Mask-out & 9.2 & 9.4 & 11.4 & 16.3 & 18.8 & 29.2 & 6.1 & 1.6 & 8.9 & 4.1 & 21.3 & 15.0 \\
\midrule
\multicolumn{13}{@{}l}{\emph{Generative}} \\
DeepPrivacy2 (body) & \best{6.6} & \best{8.5} & \best{15.3} & 6.9 & \best{12.5} & 57.3 & 33.1 & 33.4 & 30.1 & \best{4.9} & 74.5 & 63.8 \\
DeepPrivacy2 (face) & 29.4 & 35.8 & 35.7 & 14.9 & 14.6 & 76.5 & 97.5 & 97.7 & 95.2 & 22.4 & 76.6 & 84.6 \\
UniVideo & 25.0 & 25.5 & 27.5 & 20.5 & 18.8 & 66.5 & 50.2 & 61.6 & 67.0 & 53.2 & 72.3 & 61.7 \\
ReGenHuman & 12.6 & 15.5 & 33.2 & \best{6.0} & 22.2 & 66.9 & 63.8 & 77.2 & 80.8 & 26.3 & 70.2 & 79.2 \\
\method, free-text prompt & 14.5 & 13.1 & 17.6 & 19.1 & 18.8 & \best{47.4} & 27.7 & 25.2 & 25.6 & 5.1 & 53.2 & \best{49.6} \\
\method & 12.2 & 11.8 & 16.4 & 17.0 & 14.6 & \best{47.4} & \best{24.5} & \best{17.9} & \best{19.5} & \best{4.9} & \best{42.6} & 50.0 \\
\bottomrule
\end{tabular*}
\end{table*}

\begin{table*}[t]
\centering
\footnotesize
\setlength{\tabcolsep}{3pt}
\caption{\textbf{Utility and realism.} SPS: SSIM outside the person mask; $\Delta E_\mathrm{warp}$: warping error in excess of the source video; $\mathcal{A}_\mathrm{cons}$: action-logit cosine (\%); OKS and Det.: YOLOv8-pose keypoint similarity of matched persons and the share of source persons still detected (\%); FID between original and anonymized frames. Bold: best generative method.}
\label{tab:utility}
\begin{tabular*}{\textwidth}{@{\extracolsep{\fill}}lcccccccccccc}
\toprule
 & \multicolumn{6}{c}{Belvedere} & \multicolumn{6}{c}{OpenHumanVid} \\ \cmidrule(lr){2-7} \cmidrule(lr){8-13}
Method & SPS$\uparrow$ & $\Delta E_\mathrm{warp}\!\downarrow$ & $\mathcal{A}_\mathrm{cons}\!\uparrow$ & OKS$\uparrow$ & Det.$\uparrow$ & FID$\downarrow$ & SPS$\uparrow$ & $\Delta E_\mathrm{warp}\!\downarrow$ & $\mathcal{A}_\mathrm{cons}\!\uparrow$ & OKS$\uparrow$ & Det.$\uparrow$ & FID$\downarrow$ \\
\midrule
Original & -- & -- & -- & -- & -- & -- & -- & -- & -- & -- & -- & -- \\
\midrule
\multicolumn{13}{@{}l}{\emph{Heuristic}} \\
Blur & 0.992 & $-0.0003$ & 93.3 & 0.865 & 89.1 & 65.0 & 0.981 & $-0.0011$ & 55.8 & 0.628 & 81.3 & 132.8 \\
Mask-out & 0.988 & $-0.0002$ & 83.9 & 0.806 & 48.2 & 102.1 & 0.975 & $-0.0015$ & 44.8 & 0.286 & 8.1 & 219.2 \\
\midrule
\multicolumn{13}{@{}l}{\emph{Generative}} \\
DeepPrivacy2 (body) & 0.979 & 0.0028 & 92.3 & 0.888 & 93.0 & 40.4 & 0.973 & 0.0088 & 61.1 & 0.883 & 93.4 & 120.8 \\
DeepPrivacy2 (face) & \best{0.984} & 0.0014 & \best{98.6} & \best{0.951} & \best{96.1} & \best{9.9} & \best{0.984} & 0.0024 & \best{85.7} & \best{0.913} & \best{96.2} & \best{25.0} \\
UniVideo & 0.831 & \best{0.0011} & 84.5 & 0.754 & 77.4 & 65.7 & 0.852 & 0.0013 & 66.7 & 0.805 & 91.0 & 114.9 \\
ReGenHuman & 0.896 & 0.0033 & 89.9 & 0.786 & 83.6 & 62.2 & 0.939 & 0.0024 & 77.6 & 0.878 & 91.9 & 84.0 \\
\method, free-text prompt & 0.927 & 0.0028 & 90.5 & 0.627 & 83.8 & 51.3 & 0.951 & \best{0.0010} & 60.5 & 0.585 & 91.5 & 106.6 \\
\method & 0.927 & 0.0029 & 90.2 & 0.663 & 86.4 & 53.3 & 0.951 & 0.0011 & 60.7 & 0.535 & 89.5 & 113.3 \\
\bottomrule
\end{tabular*}
\end{table*}

\begin{table*}[t]
\centering
\footnotesize
\setlength{\tabcolsep}{3pt}
\caption{\textbf{Controlled ablation on Belvedere.} Each row changes one component of \method and keeps the masks, prompts, model and seed of the reference otherwise fixed. Privacy: cross-time Rank-1 (\%) and soft-biometric retention; $n$: clips generated.}
\label{tab:ablation}
\begin{tabular*}{\textwidth}{@{\extracolsep{\fill}}lccccccccccc}
\toprule
Variant & $n$ & OSNet & TransReID & CSCI & ArcFace & Attr. & SPS$\uparrow$ & $\Delta E_\mathrm{warp}\!\downarrow$ & $\mathcal{A}_\mathrm{cons}\!\uparrow$ & OKS$\uparrow$ \\
\midrule
\method & 49 & 12.2 & 11.8 & 16.4 & 17.0 & 47.4 & 0.927 & 0.0029 & 90.2 & 0.663 \\
w/o LMM filter & 49 & \best{7.4} & \best{5.3} & \best{8.8} & \best{7.8} & 51.7 & 0.919 & 0.0029 & 87.3 & 0.641 \\
w/o temporal completion & 49 & 14.2 & 14.5 & 20.4 & 19.2 & 54.2 & 0.928 & 0.0029 & 91.0 & 0.708 \\
neutral identity prompt & 49 & 12.6 & 13.9 & 21.0 & 20.7 & 62.7 & 0.928 & \best{0.0023} & 91.8 & 0.754 \\
factual identity prompt & 49 & 13.5 & 17.5 & 24.4 & 19.2 & 62.9 & 0.927 & 0.0027 & 92.3 & 0.691 \\
raw video as context & 49 & 23.3 & 25.1 & 28.3 & 24.2 & 72.7 & 0.933 & 0.0024 & \best{96.7} & \best{0.873} \\
context scale 0 & 49 & \best{1.6} & \best{1.0} & \best{2.0} & \best{2.3} & \best{37.5} & 0.519 & $-0.0037$ & 41.6 & 0.192 \\
\bottomrule
\end{tabular*}
\end{table*}

\begin{figure*}[t]
  \centering
  \includegraphics[width=0.8\textwidth]{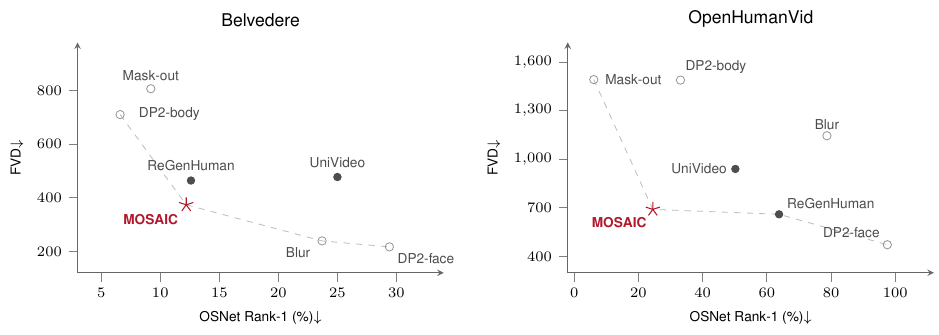}
  \caption{\textbf{Privacy and realism trade-off.} Cross-time OSNet Rank-1 against FVD. Lower is better on both axes, and the dashed line is the Pareto front. Open circles are heuristic or per-frame baselines, and filled circles are temporally coherent generative methods. \method sits at the knee. Mask-out and DeepPrivacy2 (body) reach low leakage only with about twice the FVD, and DeepPrivacy2 (face) keeps a low FVD by leaving the body identifiable. The free-text variant overlaps \method and is omitted.}
  \label{fig:tradeoff}
\end{figure*}

\Cref{tab:privacy} reports identity leakage, \cref{tab:utility} utility, and \cref{fig:tradeoff} the trade-off. Face-only anonymization leaves body Re-ID at the Original row (Belvedere OSNet 29.4, OHV 97.5). DeepPrivacy2 (body) has the lowest Belvedere appearance Rank-1 (6.6) but flickers, since it treats every frame independently (OHV $\Delta E_\mathrm{warp}$ $0.0088$ against $0.0011$ for \method). Among methods that keep a temporally coherent person, \method has the lowest cross-time Rank-1 on every OpenHumanVid attacker (OSNet 24.5, TransReID 17.9, CSCI 19.5, ArcFace 4.9, gait 42.6). Against ReGenHuman, which shares the inpainting backbone but decides neither which regions are people nor which identity to synthesize, the gap is 24.5 versus 63.8, 17.9 versus 77.2 and 42.6 versus 70.2. \method also holds the highest SPS among temporally coherent person-replacing methods (0.927 Belvedere, 0.951 OHV, against 0.831/0.852 for UniVideo, whose under-edited mask drags the background along). Gait is the price of preserved motion: the attacker reads the motion the action metrics reward. Leakage drops from 83.0 to 42.6 on OHV, the lowest among person-replacing methods, while pose-conditioned ReGenHuman keeps the signature nearly intact (70.2).

\paragraph{Protocol.}
Two protocol choices matter. First, a Re-ID-trained OSNet (MSMT17) is a far stronger attacker than the ImageNet-initialized OSNet of our earlier evaluation: on the Original row it lifts cross-time Rank-1 from 24.0 to 29.4 (Belvedere) and from 91.6 to 95.0 (OHV), and it is the weight we report in \cref{tab:privacy}. Second, cross-time retrieval is much harder than same-time on the Original row (Belvedere 100 to 29.4), which is why we use it as the main protocol.

\subsection{Ablation}

\Cref{tab:ablation} changes one component at a time on Belvedere, generated by the same runner with identical masks, prompts, model and seed. \emph{Grounding}: without the LMM filter every SAM3 candidate is regenerated, including depicted people, which is why its cross-time Rank-1 drops (the generator re-renders more of the frame, so the real person changes more) but at the cost of editing artwork, as \cref{fig:qualitative} shows. \emph{Planning}: the counterfactual identity matters most. A neutral prompt (12.6 OSNet) is close, but a factual prompt that names the source attributes raises leakage to 13.5/17.5/24.4, confirming that Stage~II must push away from the source. \emph{Conditioning} is the largest effect: feeding the raw video as context leaks the source appearance into the mask and roughly doubles Rank-1 (23.3 OSNet), while context scale 0 removes the structural hint and collapses utility (SPS 0.519, OKS 0.192) even though it is the most private row. The identity-erased context is what lets Stage~III preserve the scene.

\subsection{Qualitative results}

Mask-out and blur remove the visitors, and with them the action. On the Belvedere portrait, DeepPrivacy2 replaces the face of the painted sitter and ReGenHuman replaces the sitter (yellow boxes), while \method leaves that painting unchanged and replaces the people in front of it. On OpenHumanVid, DeepPrivacy2 turns the historical costumes into contemporary clothes, ReGenHuman keeps the dress, and \method changes the two women's clothes and faces while they remain female-presenting, leaving the corridor and their conversation in place.

\section{Conclusion}

We formalized Generative Video Anonymization, proposed a protocol with temporally disjoint retrieval and time-aligned utility, introduced the Belvedere Museum benchmark, and presented \method, a training-free pipeline of grounding, counterfactual identity planning and mask-conditioned generation. On OpenHumanVid, where the original video is re-identified at 95\% Rank-1, \method reduces appearance, clothes-changing, face and gait attackers to 24.5, 19.5, 4.9 and 42.6, with scene preservation at 0.951 SPS, the lowest leakage and the best scene preservation among temporally coherent person-replacing methods.

\paragraph{Limitations and future work.}
\label{sec:limitations}
\method inherits the resolution and clip length of its pretrained generator and processes longer videos in segments. No person-preserving anonymization erases identity. Residual signal remains in the motion we deliberately preserve, and occasionally in a source attribute echoed by the scene caption. Our attacker suite is off the shelf, and adaptive attackers are a natural next step.

\paragraph{Ethics and broader impact.}
This work aligns with institutional ethical standards and was approved by the university's ethics committee. All recordings were collected with the informed consent of the participants, and signage notified other visitors of the recording area. Data is stored locally and access is restricted to authorized project members. Publication of the Belvedere footage awaits museum approval, because it contains copyrighted artwork.

This work is a way to release human-centric video under privacy regulation. Such footage is often withheld because it identifies the people in it. Replacing the person keeps the action for downstream behaviour analysis and lowers the risk that motion alone identifies them. Drawing the new identity at random avoids mapping every source person onto one stereotype. Release still requires consent, institutional approval and access control.

\section*{LLM Usage Statement}
Large language model assistants (Cursor) were used for language editing and condensation of author-written text, for assistance with analysis code and data checking, and for aligning the bibliography with publisher records. All experiments were run with the authors' own pipeline, and every reported number, figure, and claim was checked by the authors, who take full responsibility for the content of this paper.

{
    \small
    \bibliographystyle{ieeenat_fullname}
    \bibliography{main-correct}
}

\clearpage
\appendix
\newcommand{\supptable}[1]{\input{tables/#1}}
\newenvironment{promptbox}{\begin{quote}\raggedright\footnotesize\ttfamily}{\end{quote}}

\section{Implementation details}
\label{sec:supp_impl}

\subsection{Stage I: grounding}
SAM3 proposes person instances in every frame. For each instance the LMM receives two images, the full frame with the instance highlighted and a close-up of the instance (its box expanded by 30\% of its width and height on each side), and the prompt below. We read the next-token logits of the four letters, take the arg-max of their softmax, and keep the instance if the answer is \texttt{A} or \texttt{D}; artworks and shadows are discarded, whereas uncertain instances are kept so that a missed person does not leak an identity. Kept masks are completed over frames where the track is missed by optical-flow propagation (gaps of at most 20 frames) and dilated by 6\,px.
\begin{promptbox}
You are a professional visual inspection system.\\
You are given TWO images of the SAME highlighted region:\\
- Image A: the full scene with environmental context\\
- Image B: a close-up view of the highlighted region\\[2pt]
Your task is to determine the PHYSICAL NATURE of the highlighted figure.\\[2pt]
Categories:\\
A: Real human --- a physically present person with a 3D body occupying space in the real world.\\
B: Artwork --- a person depicted in a flat artwork or painting attached to a surface (2D, no real depth).\\
C: Shadow --- a cast shadow of a human body projected onto a surface, without a physical body present.\\
D: Uncertain --- visual evidence is insufficient or ambiguous.\\[2pt]
Key visual cues:\\
- Real humans show true 3D depth, volume, self-occlusion, and consistent interaction with the environment.\\
- Artworks are flat, lack physical depth, and remain fixed on walls or surfaces.\\
- Shadows lack volume, have no surface texture, and depend on external light direction.\\[2pt]
Decision rules:\\
- Choose A if there is clear evidence of a 3D physical body.\\
- Choose B ONLY if you are confident the figure is a flat artwork.\\
- Choose C ONLY if the figure is clearly a cast shadow without a visible body.\\
- Otherwise, choose D.\\[2pt]
Output ONLY ONE letter: A, B, C, or D.
\end{promptbox}

\subsection{Stage II: counterfactual identity}
Four keyframes of the clip are described one by one (\emph{``Briefly describe this video frame. Include: 1) The scene. 2) The main person's appearance (gender, age, build) and action. Be concise.''}), the descriptions are summarized into one paragraph of at most 100 words that starts with ``A [person description]'', and the source attributes are read from that paragraph as JSON with the keys gender, age group, body type, skin tone and clothing. The target identity is then drawn by the rules of \cref{tab:supp_identity}: every attribute is moved to a different value, and the clothing is drawn from a fixed list, so the target never depends on the source appearance beyond these coarse categories. The gender entry is the label written into the prompt; the rendered person often keeps the source presentation (\cref{tab:s_attr_belvedere,tab:s_attr_ohv}).

\begin{table}[t]
\centering
\footnotesize
\setlength{\tabcolsep}{3pt}
\begin{tabular}{@{}lp{0.62\linewidth}@{}}
\toprule
Attribute & Target given the source attribute \\
\midrule
Gender & female $\to$ Male; otherwise Female \\
Age & child, teen, young $\to$ \{Middle-aged, Elderly\}; elderly, old, senior $\to$ \{Child, Teenager, Young Adult\}; otherwise \{Young Adult, Elderly\} \\
Build & slim, average, thin $\to$ \{Large, Muscular, Athletic\}; otherwise \{Slim, Average\} \\
Skin tone & any of Pale, Fair, Tan, Olive, Brown, Dark except the source category \\
Clothing & one of: a formal business suit; a casual hoodie and jeans; a leather jacket and black pants; a bright colored t-shirt and shorts; an elegant long dress; a sweater and khaki trousers; a denim jacket over a white t-shirt; a professional uniform; sportswear with a headband; a vintage flannel shirt \\
\bottomrule
\end{tabular}
\caption{\textbf{Counterfactual identity sampling} (Stage~II). Braces denote a uniform random choice.}
\label{tab:supp_identity}
\end{table}

The generation prompt combines the target identity with the scene summary:
\begin{promptbox}
An \{age\} \{gender\} person with a \{build\} build and \{skin tone\} skin, wearing \{clothing\}. The person performs exactly the same action, pose, and motion as the person in the source video. Preserve the camera viewpoint, timing, lighting, museum environment, and every non-person region exactly. Action and scene reference: \{scene summary\} Photorealistic video, coherent anatomy, temporally stable identity.
\end{promptbox}
On OpenHumanVid, ``museum environment'' is replaced by ``environment''. The ablations change only the first sentence: the \emph{factual} prompt uses the source attributes and the \emph{neutral} prompt reads ``A photorealistic person with an unspecified identity.''

\subsection{Stage III: generation}
We use Wan2.1-VACE-1.3B~\cite{jiang2025vace} at 480p ($832{\times}480$), 81 frames at 16\,fps, UniPC with 50 steps, flow shift 16, classifier-free guidance 5.0 and VACE context scale 1.0. The conditioning video is the source with the Stage~I mask region removed, and the mask video is the Stage~I mask; the reference seed is 2025 (seeds 2026 and 2027 are rerun on five hard clips, \cref{sec:supp_tables}). A clip takes a median of 409\,s (407--473\,s over the 49 Belvedere clips) on one A100 80GB.

\section{Evaluation details}
\label{sec:supp_eval}

\paragraph{Frame and region correspondence.}
\Cref{tab:supp_align} lists how each method maps output frames to source frames and which source region an output frame shows. The identity mappings hold for methods that process every frame at the source resolution. For VACE-based outputs (\method and all ablations) the input video is resampled to $\min(81, 4k{+}1)$ frames spread uniformly over the clip and resized to cover $832{\times}480$ before a center crop, which keeps 97.5\% of the width of a 16:9 frame. ReGenHuman samples every $\operatorname{round}(\mathit{fps}/7.5)$-th frame (every third at 25\,fps, at most 49) and stretches the full frame. UniVideo conditions on the first $4\lfloor(\min(n_\mathrm{out}, n)-1)/4\rfloor + 1$ source frames and resizes to an area of $320{\cdot}576$ pixels before a top-left crop to multiples of 16. How its output frames spread over that window is not fixed by the code, so we align each clip by monotone dynamic time warping of $104{\times}60$ grayscale frames on the background (Stage~I mask dilated and excluded), with a free start in the first quarter and a free end, and fall back to the frame-by-frame reading unless the warp lowers the mean matching cost by 15\%. The warp is kept for 46 of the 49 Belvedere clips and 12 of the 50 OHV clips. For the Belvedere UniVideo outputs at $848{\times}480$ the crop is estimated on four clips by a grid search over centered and top-left boxes (box $[0.006, 0.004, 0.994, 0.996]$); the estimates for the other crops agree with the formulas to within 0.4\% of the frame width.

\begin{table}[t]
\centering
\scriptsize
\setlength{\tabcolsep}{2pt}
\begin{tabular}{@{}p{0.21\linewidth}p{0.35\linewidth}p{0.35\linewidth}@{}}
\toprule
Method & Output frame $j$ shows & Source region \\
\midrule
Blur, Mask-out, DP2 (body, face) & frame $j$ & full frame \\
\method, ablations & $\operatorname{round}(j(n-1)/(n_\mathrm{out}-1))$ & cover-resize, center crop \\
ReGenHuman & $j \cdot \operatorname{round}(\mathit{fps}/7.5)$ & full frame (stretched) \\
UniVideo & DTW or $j$, within the first $4\lfloor(\min(n_\mathrm{out},n)-1)/4\rfloor+1$ & area resize, top-left crop \\
\bottomrule
\end{tabular}
\caption{\textbf{Frame and region correspondence} of each method for a source clip of $n$ frames and an output of $n_\mathrm{out}$ frames.}
\label{tab:supp_align}
\end{table}

\paragraph{Common span.}
For each clip, every metric uses source frames $[0,S]$, where $S$ is the last source frame covered by every method that has an output for the clip. On Belvedere $S=120$ for 38 clips and $S=60$ for the 11 clips where UniVideo returns 61 frames, so those 11 clips are compared on their first half; on OHV the span covers a median of 97\% of the clip. Frames outside the span are never used, which avoids rewarding a method for frames it did not generate.

\paragraph{Source cropping.}
Before any pixel-wise comparison with an output (SPS, warping error of the source, attribute crops), the source frame and its mask are cropped to the region in \cref{tab:supp_align}; person boxes detected on the source are mapped to output coordinates in the same way.

\paragraph{DeepPrivacy2 outputs.}
Three OHV sources contain an auxiliary stream whose duration is far longer than the video (\eg, 2557.9\,s for a 57-frame clip). DeepPrivacy2 reads frames until the container duration and therefore wrote the last frame again tens of thousands of times. For these clips we keep the first $n$ frames of the original output by stream copy, which we verified to be the frame-aligned anonymized clip followed by a frozen tail; new DeepPrivacy2 runs (face mode) receive the source remuxed to its video stream.

\section{Additional results}
\label{sec:supp_tables}

Two earlier prompt variants of \method appear in these tables: \emph{free-text prompt} passes Stage~II's free-form counterfactual description to the generator, and \emph{caption prompt} (OpenHumanVid only) uses the generic scene caption alone as the prompt.

\begin{table}[!tb]
\centering
\footnotesize
\setlength{\tabcolsep}{4pt}
\caption{\textbf{Cost per Belvedere clip.} SAM3, LMM verification and the one-clip generation row are medians over three hard clips on an A100 80GB shared through MPS; the last row is the median over the 49 clips of the reference run on a whole A100 80GB with model offloading. Memory is peak allocated CUDA memory; LMM calls are per clip.}
\label{tab:efficiency}
\begin{tabular}{lccc}
\toprule
Stage & Time (s) & Mem.\ (GiB) & LMM calls \\
\midrule
I: SAM3 & 123 & -- & -- \\
I: LMM verification & 111 & 8.9 & 261 \\
II: identity planning & -- & -- & 7 \\
III: generation & 617 & 19.8 & -- \\
III: generation, 49 clips & 409 & -- & -- \\
\bottomrule
\end{tabular}
\end{table}

\begin{table*}[t]
\centering
\scriptsize
\setlength{\tabcolsep}{4pt}
\caption{\textbf{Effect of the retrieval protocol and of the Re-ID weights} (Rank-1, \%). OSNet-IN is the ImageNet-initialized OSNet of our earlier evaluation, OSNet is trained on MSMT17. \emph{same}: query and gallery from the same 20 time points; \emph{cross}: first versus last 40\% of the clip.}
\label{tab:protocol}
\begin{tabular}{lcccccccc}
\toprule
 & \multicolumn{4}{c}{Belvedere} & \multicolumn{4}{c}{OpenHumanVid} \\ \cmidrule(lr){2-5} \cmidrule(lr){6-9}
 & \multicolumn{2}{c}{OSNet-IN} & \multicolumn{2}{c}{OSNet} & \multicolumn{2}{c}{OSNet-IN} & \multicolumn{2}{c}{OSNet} \\
Method & same & cross & same & cross & same & cross & same & cross \\
\midrule
Original & 100.0 & 24.0 & 100.0 & 29.4 & 100.0 & 91.6 & 100.0 & 95.0 \\
\midrule
Blur & 55.5 & 12.7 & 68.7 & 23.7 & 41.6 & 36.8 & 78.9 & 78.7 \\
DeepPrivacy2 (body) & 59.4 & 12.3 & 20.5 & 6.6 & 65.5 & 57.4 & 36.3 & 33.1 \\
UniVideo & 58.4 & 19.2 & 60.9 & 25.0 & 69.9 & 56.6 & 59.0 & 50.2 \\
ReGenHuman & 51.3 & 20.3 & 23.2 & 12.6 & 74.8 & 65.9 & 70.5 & 63.8 \\
\method & 35.1 & 14.6 & 22.5 & 12.2 & 42.0 & 34.1 & 25.9 & 24.5 \\
\bottomrule
\end{tabular}
\end{table*}

\begin{table*}[t]
\centering
\scriptsize
\setlength{\tabcolsep}{3pt}
\caption{\textbf{Rank-1 (\%) under both protocols, Belvedere.} same: query and gallery from the same 20 time points; cross: first versus last 40\% of the clip (gait: first versus second half of 40 frames).}
\label{tab:s_retrieval_belvedere}
\begin{tabular}{lcccccccccccccc}
\toprule
 & \multicolumn{2}{c}{OSNet-IN} & \multicolumn{2}{c}{OSNet} & \multicolumn{2}{c}{OSNet-AIN} & \multicolumn{2}{c}{TransReID} & \multicolumn{2}{c}{CSCI} & \multicolumn{2}{c}{ArcFace} & \multicolumn{2}{c}{Gait} \\ \cmidrule(lr){2-3} \cmidrule(lr){4-5} \cmidrule(lr){6-7} \cmidrule(lr){8-9} \cmidrule(lr){10-11} \cmidrule(lr){12-13} \cmidrule(lr){14-15}
Method & same & cross & same & cross & same & cross & same & cross & same & cross & same & cross & same & cross \\
\midrule
Original & 100.0 & 24.0 & 100.0 & 29.4 & 100.0 & 36.2 & 100.0 & 32.4 & 100.0 & 33.1 & 100.0 & 26.2 & 100.0 & 18.8 \\
\midrule
Blur & 55.5 & 12.7 & 68.7 & 23.7 & 71.3 & 29.9 & 75.4 & 26.5 & 67.9 & 27.6 & 49.0 & 16.7 & 98.0 & 22.9 \\
Mask-out & 17.6 & 8.5 & 16.7 & 9.2 & 18.9 & 11.7 & 15.7 & 9.4 & 26.9 & 11.4 & 31.6 & 16.3 & 77.6 & 18.8 \\
DeepPrivacy2 (body) & 59.4 & 12.3 & 20.5 & 6.6 & 24.6 & 9.6 & 27.6 & 8.5 & 38.9 & 15.3 & 22.2 & 6.9 & 93.9 & 12.5 \\
DeepPrivacy2 (face) & 96.9 & 24.1 & 97.8 & 29.4 & 98.3 & 35.5 & 98.2 & 35.8 & 93.2 & 35.7 & 40.4 & 14.9 & 100.0 & 14.6 \\
UniVideo & 58.4 & 19.2 & 60.9 & 25.0 & 63.1 & 25.6 & 63.7 & 25.5 & 59.3 & 27.5 & 45.1 & 20.5 & 85.7 & 18.8 \\
ReGenHuman & 51.3 & 20.3 & 23.2 & 12.6 & 31.1 & 20.0 & 31.2 & 15.5 & 46.6 & 33.2 & 9.9 & 6.0 & 97.3 & 22.2 \\
\method, free-text prompt & 33.7 & 12.0 & 23.1 & 14.5 & 21.9 & 13.4 & 22.4 & 13.1 & 25.8 & 17.6 & 27.4 & 19.1 & 89.8 & 18.8 \\
\method & 35.1 & 14.6 & 22.5 & 12.2 & 22.9 & 12.4 & 21.1 & 11.8 & 24.6 & 16.4 & 27.2 & 17.0 & 89.8 & 14.6 \\
\bottomrule
\end{tabular}
\end{table*}

\begin{table*}[t]
\centering
\scriptsize
\setlength{\tabcolsep}{3pt}
\caption{\textbf{Rank-1 (\%) under both protocols, OpenHumanVid.} same: query and gallery from the same 20 time points; cross: first versus last 40\% of the clip (gait: first versus second half of 40 frames).}
\label{tab:s_retrieval_ohv}
\begin{tabular}{lcccccccccccccc}
\toprule
 & \multicolumn{2}{c}{OSNet-IN} & \multicolumn{2}{c}{OSNet} & \multicolumn{2}{c}{OSNet-AIN} & \multicolumn{2}{c}{TransReID} & \multicolumn{2}{c}{CSCI} & \multicolumn{2}{c}{ArcFace} & \multicolumn{2}{c}{Gait} \\ \cmidrule(lr){2-3} \cmidrule(lr){4-5} \cmidrule(lr){6-7} \cmidrule(lr){8-9} \cmidrule(lr){10-11} \cmidrule(lr){12-13} \cmidrule(lr){14-15}
Method & same & cross & same & cross & same & cross & same & cross & same & cross & same & cross & same & cross \\
\midrule
Original & 100.0 & 91.6 & 100.0 & 95.0 & 100.0 & 94.3 & 100.0 & 96.0 & 100.0 & 96.2 & 100.0 & 98.4 & 100.0 & 83.0 \\
\midrule
Blur & 41.6 & 36.8 & 78.9 & 78.7 & 79.6 & 77.5 & 82.5 & 80.4 & 84.1 & 85.1 & 71.9 & 68.2 & 77.6 & 61.7 \\
Mask-out & 9.7 & 8.9 & 9.6 & 6.1 & 11.5 & 4.6 & 6.9 & 1.6 & 10.2 & 8.9 & 6.0 & 4.1 & 32.7 & 21.3 \\
DeepPrivacy2 (body) & 65.5 & 57.4 & 36.3 & 33.1 & 51.9 & 48.2 & 33.5 & 33.4 & 32.8 & 30.1 & 10.4 & 4.9 & 83.7 & 74.5 \\
DeepPrivacy2 (face) & 99.8 & 93.1 & 99.9 & 97.5 & 100.0 & 96.3 & 99.9 & 97.7 & 97.1 & 95.2 & 24.7 & 22.4 & 93.9 & 76.6 \\
UniVideo & 69.9 & 56.6 & 59.0 & 50.2 & 69.2 & 64.5 & 67.8 & 61.6 & 69.6 & 67.0 & 56.2 & 53.2 & 87.8 & 72.3 \\
ReGenHuman & 74.8 & 65.9 & 70.5 & 63.8 & 78.0 & 73.7 & 77.6 & 77.2 & 84.1 & 80.8 & 24.5 & 26.3 & 85.7 & 70.2 \\
\method, free-text prompt & 42.8 & 32.5 & 33.3 & 27.7 & 39.4 & 35.5 & 30.8 & 25.2 & 27.6 & 25.6 & 7.6 & 5.1 & 63.3 & 53.2 \\
\method, caption prompt & 45.8 & 42.6 & 35.3 & 32.4 & 35.7 & 31.4 & 26.3 & 26.5 & 40.2 & 38.8 & 11.1 & 7.7 & 74.4 & 51.2 \\
\method & 42.0 & 34.1 & 25.9 & 24.5 & 33.8 & 33.5 & 19.5 & 17.9 & 22.4 & 19.5 & 7.0 & 4.9 & 67.3 & 42.6 \\
\bottomrule
\end{tabular}
\end{table*}

\begin{table*}[t]
\centering
\tiny
\setlength{\tabcolsep}{3pt}
\caption{\textbf{Cross-time mAP (\%), identity similarity $\mathcal{S}_{\cos}$ and number of gallery identities $N$, Belvedere.} $N$ drops when a method makes persons, faces or silhouettes undetectable; their queries then count as misses.}
\label{tab:s_map_belvedere}
\begin{tabular}{lccccccccccccccccccccc}
\toprule
 & \multicolumn{3}{c}{OSNet-IN} & \multicolumn{3}{c}{OSNet} & \multicolumn{3}{c}{OSNet-AIN} & \multicolumn{3}{c}{TransReID} & \multicolumn{3}{c}{CSCI} & \multicolumn{3}{c}{ArcFace} & \multicolumn{3}{c}{Gait} \\ \cmidrule(lr){2-4} \cmidrule(lr){5-7} \cmidrule(lr){8-10} \cmidrule(lr){11-13} \cmidrule(lr){14-16} \cmidrule(lr){17-19} \cmidrule(lr){20-22}
Method & mAP & $\mathcal{S}_{\cos}$ & $N$ & mAP & $\mathcal{S}_{\cos}$ & $N$ & mAP & $\mathcal{S}_{\cos}$ & $N$ & mAP & $\mathcal{S}_{\cos}$ & $N$ & mAP & $\mathcal{S}_{\cos}$ & $N$ & mAP & $\mathcal{S}_{\cos}$ & $N$ & mAP & $\mathcal{S}_{\cos}$ & $N$ \\
\midrule
Original & 12.0 & 0.708 & 48 & 16.7 & 0.555 & 48 & 18.8 & 0.611 & 48 & 17.7 & 0.570 & 48 & 18.2 & 0.566 & 48 & 19.2 & 0.348 & 44 & 30.1 & 1.000 & 47 \\
\midrule
Blur & 9.3 & 0.679 & 48 & 13.9 & 0.508 & 48 & 16.8 & 0.570 & 48 & 15.4 & 0.531 & 48 & 17.2 & 0.539 & 48 & 16.0 & 0.264 & 38 & 31.4 & 0.963 & 47 \\
Mask-out & 9.1 & 0.643 & 33 & 9.3 & 0.425 & 33 & 10.2 & 0.512 & 33 & 9.4 & 0.416 & 33 & 10.3 & 0.465 & 33 & 15.2 & 0.252 & 26 & 26.9 & 0.942 & 44 \\
DeepPrivacy2 (body) & 7.1 & 0.669 & 48 & 4.8 & 0.382 & 48 & 5.6 & 0.475 & 48 & 5.8 & 0.424 & 48 & 8.3 & 0.449 & 48 & 6.2 & 0.115 & 45 & 23.8 & 0.962 & 47 \\
DeepPrivacy2 (face) & 11.5 & 0.701 & 48 & 16.2 & 0.541 & 48 & 18.0 & 0.598 & 48 & 17.5 & 0.556 & 48 & 17.4 & 0.539 & 48 & 7.5 & 0.124 & 49 & 27.8 & 0.978 & 47 \\
UniVideo & 11.0 & 0.666 & 48 & 13.8 & 0.475 & 48 & 13.6 & 0.541 & 48 & 14.3 & 0.497 & 48 & 14.4 & 0.496 & 48 & 13.9 & 0.158 & 42 & 30.1 & 0.953 & 47 \\
ReGenHuman & 12.2 & 0.668 & 37 & 10.1 & 0.417 & 37 & 12.8 & 0.503 & 37 & 12.4 & 0.451 & 37 & 20.5 & 0.477 & 37 & 7.9 & 0.111 & 35 & 33.4 & 0.955 & 37 \\
\method, free-text prompt & 7.3 & 0.657 & 48 & 8.5 & 0.426 & 48 & 9.2 & 0.500 & 48 & 7.5 & 0.433 & 48 & 11.7 & 0.417 & 48 & 10.7 & 0.118 & 46 & 29.1 & 0.947 & 47 \\
\method & 8.0 & 0.651 & 48 & 8.1 & 0.424 & 48 & 8.9 & 0.497 & 48 & 7.9 & 0.431 & 48 & 10.1 & 0.416 & 48 & 10.9 & 0.128 & 45 & 26.6 & 0.948 & 47 \\
\bottomrule
\end{tabular}
\end{table*}

\begin{table*}[t]
\centering
\tiny
\setlength{\tabcolsep}{3pt}
\caption{\textbf{Cross-time mAP (\%), identity similarity $\mathcal{S}_{\cos}$ and number of gallery identities $N$, OpenHumanVid.} $N$ drops when a method makes persons, faces or silhouettes undetectable; their queries then count as misses.}
\label{tab:s_map_ohv}
\begin{tabular}{lccccccccccccccccccccc}
\toprule
 & \multicolumn{3}{c}{OSNet-IN} & \multicolumn{3}{c}{OSNet} & \multicolumn{3}{c}{OSNet-AIN} & \multicolumn{3}{c}{TransReID} & \multicolumn{3}{c}{CSCI} & \multicolumn{3}{c}{ArcFace} & \multicolumn{3}{c}{Gait} \\ \cmidrule(lr){2-4} \cmidrule(lr){5-7} \cmidrule(lr){8-10} \cmidrule(lr){11-13} \cmidrule(lr){14-16} \cmidrule(lr){17-19} \cmidrule(lr){20-22}
Method & mAP & $\mathcal{S}_{\cos}$ & $N$ & mAP & $\mathcal{S}_{\cos}$ & $N$ & mAP & $\mathcal{S}_{\cos}$ & $N$ & mAP & $\mathcal{S}_{\cos}$ & $N$ & mAP & $\mathcal{S}_{\cos}$ & $N$ & mAP & $\mathcal{S}_{\cos}$ & $N$ & mAP & $\mathcal{S}_{\cos}$ & $N$ \\
\midrule
Original & 76.8 & 0.857 & 48 & 81.2 & 0.824 & 48 & 81.2 & 0.851 & 48 & 84.2 & 0.846 & 48 & 87.7 & 0.846 & 48 & 83.7 & 0.688 & 49 & 88.2 & 1.000 & 49 \\
\midrule
Blur & 34.5 & 0.653 & 44 & 71.2 & 0.644 & 44 & 70.3 & 0.700 & 44 & 73.2 & 0.658 & 44 & 77.0 & 0.695 & 44 & 60.4 & 0.319 & 42 & 73.0 & 0.914 & 47 \\
Mask-out & 9.9 & 0.580 & 8 & 8.7 & 0.440 & 8 & 6.5 & 0.513 & 8 & 4.4 & 0.313 & 8 & 10.4 & 0.336 & 8 & 4.9 & 0.093 & 3 & 27.3 & 0.842 & 23 \\
DeepPrivacy2 (body) & 44.6 & 0.701 & 49 & 26.8 & 0.489 & 49 & 39.3 & 0.573 & 49 & 27.4 & 0.432 & 49 & 27.2 & 0.370 & 49 & 6.8 & 0.026 & 49 & 80.3 & 0.921 & 50 \\
DeepPrivacy2 (face) & 76.9 & 0.820 & 49 & 81.8 & 0.783 & 49 & 83.3 & 0.816 & 49 & 85.4 & 0.798 & 49 & 80.6 & 0.639 & 49 & 13.2 & 0.048 & 49 & 84.0 & 0.953 & 49 \\
UniVideo & 50.4 & 0.712 & 48 & 45.4 & 0.566 & 48 & 57.1 & 0.653 & 48 & 56.3 & 0.553 & 48 & 63.6 & 0.581 & 48 & 50.7 & 0.225 & 48 & 79.3 & 0.909 & 49 \\
ReGenHuman & 55.1 & 0.720 & 47 & 59.4 & 0.607 & 47 & 65.0 & 0.672 & 47 & 68.1 & 0.615 & 47 & 71.7 & 0.639 & 47 & 23.8 & 0.085 & 49 & 78.5 & 0.918 & 47 \\
\method, free-text prompt & 30.6 & 0.685 & 49 & 30.7 & 0.500 & 49 & 34.4 & 0.582 & 49 & 25.7 & 0.432 & 49 & 24.6 & 0.377 & 49 & 5.8 & 0.024 & 49 & 66.3 & 0.876 & 49 \\
\method, caption prompt & 38.8 & 0.688 & 42 & 36.9 & 0.520 & 42 & 32.9 & 0.571 & 42 & 30.3 & 0.460 & 42 & 38.1 & 0.479 & 42 & 10.4 & 0.033 & 41 & 67.3 & 0.898 & 43 \\
\method & 28.3 & 0.669 & 46 & 24.1 & 0.469 & 46 & 30.5 & 0.548 & 46 & 21.2 & 0.396 & 46 & 20.8 & 0.391 & 46 & 7.5 & 0.025 & 48 & 55.0 & 0.866 & 48 \\
\bottomrule
\end{tabular}
\end{table*}

\begin{table*}[t]
\centering
\scriptsize
\setlength{\tabcolsep}{3pt}
\caption{\textbf{95\% bootstrap confidence intervals over clips, Belvedere} (cross-time Rank-1 in \%, 10{,}000 resamples).}
\label{tab:s_ci_belvedere}
\begin{tabular}{lccccccc}
\toprule
Method & OSNet & TransReID & CSCI & ArcFace & SPS$\uparrow$ & $\Delta E_\mathrm{warp}\!\downarrow$ & $\mathcal{A}_\mathrm{cons}\!\uparrow$ \\
\midrule
Original & 29.4 {\tiny[21.2, 37.9]} & 32.4 {\tiny[23.4, 41.6]} & 33.1 {\tiny[24.2, 42.1]} & 26.2 {\tiny[18.8, 34.2]} & -- & -- & -- \\
\midrule
Blur & 23.7 {\tiny[16.1, 31.8]} & 26.5 {\tiny[19.1, 34.3]} & 27.6 {\tiny[19.3, 36.5]} & 16.7 {\tiny[10.2, 23.9]} & 0.992 {\tiny[0.992, 0.992]} & -0.0003 {\tiny[-0.0004, -0.0003]} & 93.3 {\tiny[91.8, 94.6]} \\
Mask-out & 9.2 {\tiny[3.3, 16.5]} & 9.4 {\tiny[4.0, 15.8]} & 11.4 {\tiny[6.0, 17.9]} & 16.3 {\tiny[9.5, 23.9]} & 0.988 {\tiny[0.988, 0.989]} & -0.0002 {\tiny[-0.0004, -0.0000]} & 83.9 {\tiny[80.7, 86.8]} \\
DeepPrivacy2 (body) & 6.6 {\tiny[3.7, 9.9]} & 8.5 {\tiny[4.7, 13.0]} & 15.3 {\tiny[8.6, 23.2]} & 6.9 {\tiny[3.0, 11.9]} & 0.979 {\tiny[0.977, 0.982]} & 0.0028 {\tiny[0.0023, 0.0034]} & 92.3 {\tiny[90.5, 94.0]} \\
DeepPrivacy2 (face) & 29.4 {\tiny[21.3, 37.8]} & 35.8 {\tiny[26.5, 45.4]} & 35.7 {\tiny[26.2, 45.4]} & 14.9 {\tiny[8.5, 21.9]} & 0.984 {\tiny[0.981, 0.986]} & 0.0014 {\tiny[0.0009, 0.0020]} & 98.6 {\tiny[98.1, 99.1]} \\
UniVideo & 25.0 {\tiny[17.0, 33.4]} & 25.5 {\tiny[17.6, 34.0]} & 27.5 {\tiny[18.1, 37.2]} & 20.5 {\tiny[12.4, 29.3]} & 0.831 {\tiny[0.807, 0.852]} & 0.0011 {\tiny[0.0006, 0.0015]} & 84.5 {\tiny[81.0, 87.7]} \\
ReGenHuman & 12.6 {\tiny[7.8, 18.1]} & 15.5 {\tiny[9.1, 22.7]} & 33.2 {\tiny[23.0, 43.7]} & 6.0 {\tiny[2.1, 10.8]} & 0.896 {\tiny[0.888, 0.904]} & 0.0033 {\tiny[0.0027, 0.0040]} & 89.9 {\tiny[88.3, 91.4]} \\
\method, free-text prompt & 14.5 {\tiny[8.2, 21.6]} & 13.1 {\tiny[7.9, 19.1]} & 17.6 {\tiny[11.4, 24.3]} & 19.1 {\tiny[10.7, 28.5]} & 0.927 {\tiny[0.921, 0.932]} & 0.0028 {\tiny[0.0025, 0.0032]} & 90.5 {\tiny[88.9, 92.0]} \\
\method & 12.2 {\tiny[6.0, 19.4]} & 11.8 {\tiny[6.2, 18.7]} & 16.4 {\tiny[9.5, 24.3]} & 17.0 {\tiny[10.0, 24.9]} & 0.927 {\tiny[0.922, 0.933]} & 0.0029 {\tiny[0.0025, 0.0032]} & 90.2 {\tiny[88.2, 92.0]} \\
\bottomrule
\end{tabular}
\end{table*}

\begin{table*}[t]
\centering
\scriptsize
\setlength{\tabcolsep}{3pt}
\caption{\textbf{95\% bootstrap confidence intervals over clips, OpenHumanVid} (cross-time Rank-1 in \%, 10{,}000 resamples).}
\label{tab:s_ci_ohv}
\begin{tabular}{lccccccc}
\toprule
Method & OSNet & TransReID & CSCI & ArcFace & SPS$\uparrow$ & $\Delta E_\mathrm{warp}\!\downarrow$ & $\mathcal{A}_\mathrm{cons}\!\uparrow$ \\
\midrule
Original & 95.0 {\tiny[89.2, 99.3]} & 96.0 {\tiny[90.6, 99.7]} & 96.2 {\tiny[91.2, 99.3]} & 98.4 {\tiny[96.2, 100.0]} & -- & -- & -- \\
\midrule
Blur & 78.7 {\tiny[69.1, 87.3]} & 80.4 {\tiny[70.3, 89.2]} & 85.1 {\tiny[75.9, 92.9]} & 68.2 {\tiny[55.7, 80.0]} & 0.981 {\tiny[0.976, 0.985]} & -0.0011 {\tiny[-0.0013, -0.0008]} & 55.8 {\tiny[50.5, 60.9]} \\
Mask-out & 6.1 {\tiny[1.4, 12.1]} & 1.6 {\tiny[0.0, 4.7]} & 8.9 {\tiny[2.7, 16.9]} & 4.1 {\tiny[0.0, 9.4]} & 0.975 {\tiny[0.969, 0.981]} & -0.0015 {\tiny[-0.0019, -0.0012]} & 44.8 {\tiny[38.7, 51.0]} \\
DeepPrivacy2 (body) & 33.1 {\tiny[22.7, 44.1]} & 33.4 {\tiny[22.2, 45.1]} & 30.1 {\tiny[19.1, 41.9]} & 4.9 {\tiny[2.1, 8.4]} & 0.973 {\tiny[0.965, 0.979]} & 0.0088 {\tiny[0.0072, 0.0106]} & 61.1 {\tiny[55.3, 66.7]} \\
DeepPrivacy2 (face) & 97.5 {\tiny[95.4, 99.2]} & 97.7 {\tiny[93.9, 99.9]} & 95.2 {\tiny[90.1, 98.8]} & 22.4 {\tiny[14.4, 31.0]} & 0.984 {\tiny[0.977, 0.989]} & 0.0024 {\tiny[0.0018, 0.0031]} & 85.7 {\tiny[82.0, 89.1]} \\
UniVideo & 50.2 {\tiny[37.3, 63.2]} & 61.6 {\tiny[48.7, 74.0]} & 67.0 {\tiny[54.8, 78.8]} & 53.2 {\tiny[40.4, 66.3]} & 0.852 {\tiny[0.817, 0.884]} & 0.0013 {\tiny[0.0008, 0.0019]} & 66.7 {\tiny[61.8, 71.5]} \\
ReGenHuman & 63.8 {\tiny[52.1, 74.6]} & 77.2 {\tiny[67.1, 86.5]} & 80.8 {\tiny[71.5, 89.0]} & 26.3 {\tiny[16.4, 37.0]} & 0.939 {\tiny[0.915, 0.959]} & 0.0024 {\tiny[0.0017, 0.0031]} & 77.6 {\tiny[74.1, 80.9]} \\
\method, free-text prompt & 27.7 {\tiny[17.9, 37.7]} & 25.2 {\tiny[15.6, 35.5]} & 25.6 {\tiny[15.1, 37.1]} & 5.1 {\tiny[2.0, 8.9]} & 0.951 {\tiny[0.936, 0.963]} & 0.0010 {\tiny[0.0006, 0.0014]} & 60.5 {\tiny[54.7, 66.1]} \\
\method, caption prompt & 32.4 {\tiny[21.4, 44.4]} & 26.5 {\tiny[15.1, 38.9]} & 38.8 {\tiny[26.3, 51.6]} & 7.7 {\tiny[2.6, 14.2]} & 0.951 {\tiny[0.935, 0.964]} & 0.0005 {\tiny[0.0001, 0.0010]} & 67.4 {\tiny[62.6, 72.2]} \\
\method & 24.5 {\tiny[14.8, 35.2]} & 17.9 {\tiny[9.1, 27.6]} & 19.5 {\tiny[10.0, 30.0]} & 4.9 {\tiny[1.6, 9.5]} & 0.951 {\tiny[0.937, 0.964]} & 0.0011 {\tiny[0.0006, 0.0016]} & 60.7 {\tiny[54.8, 66.2]} \\
\bottomrule
\end{tabular}
\end{table*}

\begin{table*}[t]
\centering
\scriptsize
\setlength{\tabcolsep}{3pt}
\caption{\textbf{All utility measures, Belvedere.} SPS$_\mathrm{first20}$ compares the first 20 decoded frames of each video, regardless of frame rate and crop (no alignment). $E_\mathrm{warp}(\hat V; F_V)$ warps the anonymized frames with the flow of the source; TWE$_\mathrm{flow}$ is the mean squared difference between source and anonymized flows. Top-1: action label unchanged (\%); Top-5: overlap of the top-5 labels (\%).}
\label{tab:s_utility_belvedere}
\begin{tabular}{lcccccccccccc}
\toprule
Method & SPS & SPS$_\mathrm{first20}$ & $E_\mathrm{warp}(\hat V)$ & $E_\mathrm{warp}(\hat V; F_V)$ & $E_\mathrm{warp}(V)$ & $\Delta E_\mathrm{warp}$ & TWE$_\mathrm{flow}$ & $\mathcal{A}_\mathrm{cons}$ & Top-1 & Top-5 & OKS & Det. \\
\midrule
Original & -- & -- & -- & -- & -- & -- & -- & -- & -- & -- & -- & -- \\
\midrule
Blur & 0.992 & 0.992 & 0.0049 & 0.0051 & 0.0052 & -0.0003 & 2.48 & 93.3 & 40.8 & 61.6 & 0.865 & 89.1 \\
Mask-out & 0.988 & 0.988 & 0.0050 & 0.0067 & 0.0052 & -0.0002 & 10.00 & 83.9 & 20.4 & 40.8 & 0.806 & 48.2 \\
DeepPrivacy2 (body) & 0.979 & 0.980 & 0.0081 & 0.0104 & 0.0052 & 0.0028 & 10.55 & 92.3 & 44.9 & 56.3 & 0.888 & 93.0 \\
DeepPrivacy2 (face) & 0.984 & 0.985 & 0.0066 & 0.0083 & 0.0052 & 0.0014 & 4.89 & 98.6 & 75.5 & 80.8 & 0.951 & 96.1 \\
UniVideo & 0.831 & 0.787 & 0.0066 & 0.0184 & 0.0055 & 0.0011 & 80.93 & 84.5 & 30.6 & 41.6 & 0.754 & 77.3 \\
ReGenHuman & 0.896 & 0.744 & 0.0133 & 0.0172 & 0.0100 & 0.0033 & 73.00 & 89.9 & 35.1 & 51.4 & 0.786 & 83.6 \\
\method, free-text prompt & 0.927 & 0.807 & 0.0099 & 0.0201 & 0.0070 & 0.0028 & 73.50 & 90.5 & 42.9 & 51.4 & 0.627 & 83.8 \\
\method & 0.927 & 0.808 & 0.0099 & 0.0198 & 0.0070 & 0.0029 & 74.64 & 90.2 & 36.7 & 51.8 & 0.663 & 86.4 \\
\bottomrule
\end{tabular}
\end{table*}

\begin{table*}[t]
\centering
\scriptsize
\setlength{\tabcolsep}{3pt}
\caption{\textbf{All utility measures, OpenHumanVid.} SPS$_\mathrm{first20}$ compares the first 20 decoded frames of each video, regardless of frame rate and crop (no alignment). $E_\mathrm{warp}(\hat V; F_V)$ warps the anonymized frames with the flow of the source; TWE$_\mathrm{flow}$ is the mean squared difference between source and anonymized flows. Top-1: action label unchanged (\%); Top-5: overlap of the top-5 labels (\%).}
\label{tab:s_utility_ohv}
\begin{tabular}{lcccccccccccc}
\toprule
Method & SPS & SPS$_\mathrm{first20}$ & $E_\mathrm{warp}(\hat V)$ & $E_\mathrm{warp}(\hat V; F_V)$ & $E_\mathrm{warp}(V)$ & $\Delta E_\mathrm{warp}$ & TWE$_\mathrm{flow}$ & $\mathcal{A}_\mathrm{cons}$ & Top-1 & Top-5 & OKS & Det. \\
\midrule
Original & -- & -- & -- & -- & -- & -- & -- & -- & -- & -- & -- & -- \\
\midrule
Blur & 0.981 & 0.981 & 0.0032 & 0.0033 & 0.0042 & -0.0011 & 1.27 & 55.8 & 14.0 & 20.8 & 0.628 & 81.3 \\
Mask-out & 0.975 & 0.975 & 0.0027 & 0.0029 & 0.0042 & -0.0015 & 4.22 & 44.8 & 12.0 & 13.6 & 0.286 & 8.1 \\
DeepPrivacy2 (body) & 0.973 & 0.971 & 0.0131 & 0.0243 & 0.0042 & 0.0088 & 17.37 & 61.1 & 8.0 & 19.6 & 0.883 & 93.4 \\
DeepPrivacy2 (face) & 0.984 & 0.986 & 0.0066 & 0.0090 & 0.0042 & 0.0024 & 3.53 & 85.7 & 40.0 & 48.4 & 0.913 & 96.2 \\
UniVideo & 0.852 & 0.777 & 0.0053 & 0.0115 & 0.0040 & 0.0013 & 3.14 & 66.7 & 14.0 & 28.0 & 0.805 & 91.0 \\
ReGenHuman & 0.939 & 0.857 & 0.0089 & 0.0143 & 0.0065 & 0.0024 & 15.32 & 77.6 & 26.0 & 34.8 & 0.878 & 91.9 \\
\method, free-text prompt & 0.951 & 0.865 & 0.0058 & 0.0098 & 0.0049 & 0.0010 & 6.73 & 60.5 & 10.0 & 23.2 & 0.585 & 91.5 \\
\method, caption prompt & 0.951 & 0.860 & 0.0053 & 0.0089 & 0.0048 & 0.0005 & 6.69 & 67.4 & 22.7 & 30.5 & 0.578 & 89.2 \\
\method & 0.951 & 0.865 & 0.0060 & 0.0102 & 0.0049 & 0.0011 & 6.05 & 60.7 & 12.0 & 24.8 & 0.535 & 89.5 \\
\bottomrule
\end{tabular}
\end{table*}

\begin{table*}[t]
\centering
\small
\setlength{\tabcolsep}{3pt}
\caption{\textbf{Realism, Belvedere.} Frame FID against the original frames and FVD on I3D Kinetics-400 logits, on the common 81-frame, 16\,fps, $832{\times}480$ timeline.}
\label{tab:s_realism_belvedere}
\begin{tabular}{lcc}
\toprule
Method & FID$\downarrow$ & FVD$\downarrow$ \\
\midrule
Original & -- & -- \\
\midrule
Blur & 65.0 & 239.1 \\
Mask-out & 102.1 & 807.2 \\
DeepPrivacy2 (body) & 40.4 & 710.8 \\
DeepPrivacy2 (face) & 9.9 & 216.7 \\
UniVideo & 65.7 & 477.6 \\
ReGenHuman & 62.2 & 464.5 \\
\method, free-text prompt & 51.3 & 352.7 \\
\method & 53.3 & 373.9 \\
\bottomrule
\end{tabular}
\end{table*}

\begin{table*}[t]
\centering
\small
\setlength{\tabcolsep}{3pt}
\caption{\textbf{Realism, OpenHumanVid.} Frame FID against the original frames and FVD on I3D Kinetics-400 logits, on the common 81-frame, 16\,fps, $832{\times}480$ timeline.}
\label{tab:s_realism_ohv}
\begin{tabular}{lcc}
\toprule
Method & FID$\downarrow$ & FVD$\downarrow$ \\
\midrule
Original & -- & -- \\
\midrule
Blur & 132.8 & 1144.9 \\
Mask-out & 219.2 & 1491.1 \\
DeepPrivacy2 (body) & 120.8 & 1488.0 \\
DeepPrivacy2 (face) & 25.0 & 473.0 \\
UniVideo & 114.9 & 940.2 \\
ReGenHuman & 84.0 & 660.6 \\
\method, free-text prompt & 106.6 & 717.1 \\
\method, caption prompt & 104.3 & 749.3 \\
\method & 113.3 & 690.9 \\
\bottomrule
\end{tabular}
\end{table*}

\begin{table*}[t]
\centering
\small
\setlength{\tabcolsep}{3pt}
\caption{\textbf{Soft-biometric retention, Belvedere} (\%): share of clips whose anonymized label equals the original one (ret.), and the agreement expected from the two label marginals (chance).}
\label{tab:s_attr_belvedere}
\begin{tabular}{lcccccccccc}
\toprule
 & \multicolumn{2}{c}{Gender} & \multicolumn{2}{c}{Age} & \multicolumn{2}{c}{Upper} & \multicolumn{2}{c}{Lower} & \multicolumn{2}{c}{Hair} \\ \cmidrule(lr){2-3} \cmidrule(lr){4-5} \cmidrule(lr){6-7} \cmidrule(lr){8-9} \cmidrule(lr){10-11}
Method & ret. & chance & ret. & chance & ret. & chance & ret. & chance & ret. & chance \\
\midrule
Original & -- & -- & -- & -- & -- & -- & -- & -- & -- & -- \\
\midrule
Blur & 28.6 & 20.8 & 46.2 & 45.0 & 40.9 & 33.5 & 24.3 & 13.7 & 19.4 & 16.4 \\
Mask-out & 2.9 & 2.9 & 25.6 & 26.0 & 68.2 & 63.1 & 29.7 & 26.7 & 19.4 & 17.3 \\
DeepPrivacy2 (body) & 74.3 & 52.2 & 82.1 & 82.8 & 59.1 & 63.1 & 45.9 & 35.8 & 25.0 & 28.6 \\
DeepPrivacy2 (face) & 88.6 & 54.3 & 84.6 & 76.1 & 72.7 & 53.1 & 75.7 & 38.1 & 61.1 & 28.8 \\
UniVideo & 74.3 & 53.1 & 87.2 & 83.4 & 61.4 & 48.6 & 59.5 & 42.1 & 50.0 & 35.9 \\
ReGenHuman & 92.6 & 51.0 & 93.5 & 87.7 & 60.6 & 47.8 & 41.4 & 39.5 & 46.4 & 34.8 \\
\method, free-text prompt & 25.7 & 39.6 & 89.7 & 85.6 & 47.7 & 47.1 & 37.8 & 38.3 & 36.1 & 34.0 \\
\method & 45.7 & 42.4 & 82.1 & 83.4 & 40.9 & 49.0 & 32.4 & 34.1 & 36.1 & 32.1 \\
\bottomrule
\end{tabular}
\end{table*}

\begin{table*}[t]
\centering
\small
\setlength{\tabcolsep}{3pt}
\caption{\textbf{Soft-biometric retention, OpenHumanVid} (\%): share of clips whose anonymized label equals the original one (ret.), and the agreement expected from the two label marginals (chance).}
\label{tab:s_attr_ohv}
\begin{tabular}{lcccccccccc}
\toprule
 & \multicolumn{2}{c}{Gender} & \multicolumn{2}{c}{Age} & \multicolumn{2}{c}{Upper} & \multicolumn{2}{c}{Lower} & \multicolumn{2}{c}{Hair} \\ \cmidrule(lr){2-3} \cmidrule(lr){4-5} \cmidrule(lr){6-7} \cmidrule(lr){8-9} \cmidrule(lr){10-11}
Method & ret. & chance & ret. & chance & ret. & chance & ret. & chance & ret. & chance \\
\midrule
Original & -- & -- & -- & -- & -- & -- & -- & -- & -- & -- \\
\midrule
Blur & 47.9 & 25.8 & 68.8 & 66.0 & 52.1 & 34.9 & 37.5 & 16.1 & 37.5 & 25.3 \\
Mask-out & 0.0 & 0.0 & 14.6 & 14.0 & 43.8 & 40.3 & 10.4 & 13.7 & 6.2 & 8.2 \\
DeepPrivacy2 (body) & 83.3 & 53.1 & 93.8 & 93.9 & 41.7 & 41.4 & 50.0 & 29.3 & 50.0 & 29.9 \\
DeepPrivacy2 (face) & 87.5 & 50.3 & 91.7 & 88.0 & 93.8 & 36.8 & 72.9 & 25.7 & 77.1 & 33.9 \\
UniVideo & 87.5 & 47.9 & 75.0 & 76.2 & 43.8 & 27.7 & 47.9 & 24.6 & 54.2 & 27.3 \\
ReGenHuman & 95.8 & 53.1 & 93.8 & 89.9 & 70.8 & 30.0 & 54.2 & 24.4 & 81.2 & 37.1 \\
\method, free-text prompt & 29.2 & 46.9 & 89.6 & 83.9 & 47.9 & 35.7 & 31.2 & 25.7 & 50.0 & 34.8 \\
\method, caption prompt & 81.0 & 50.1 & 88.1 & 88.5 & 33.3 & 23.6 & 47.6 & 30.1 & 50.0 & 28.6 \\
\method & 35.4 & 44.3 & 77.1 & 74.0 & 52.1 & 43.7 & 35.4 & 24.7 & 50.0 & 36.5 \\
\bottomrule
\end{tabular}
\end{table*}

\begin{table*}[t]
\centering
\small
\caption{\textbf{Generation seed} on the 5 predeclared hard Belvedere clips: mean over clips, and the per-clip standard deviation across the three seeds averaged over clips.}
\label{tab:s_seeds}
\begin{tabular}{lccccccc}
\toprule
Seed & SPS & $\Delta E_\mathrm{warp}$ & $\mathcal{A}_\mathrm{cons}$ & OKS & $\mathcal{S}_{\cos}$ OSNet & $\mathcal{S}_{\cos}$ TransReID & $\mathcal{S}_{\cos}$ CSCI \\
\midrule
2025 (reference) & 0.924 & 0.0027 & 89.8 & 0.690 & 0.427 & 0.446 & 0.428 \\
2026 & 0.925 & 0.0028 & 87.3 & 0.744 & 0.432 & 0.444 & 0.428 \\
2027 & 0.924 & 0.0028 & 88.6 & 0.763 & 0.435 & 0.449 & 0.428 \\
\midrule
std.\ across seeds & 0.001 & 0.0001 & 1.7 & 0.061 & 0.011 & 0.011 & 0.015 \\
\bottomrule
\end{tabular}
\end{table*}

\begin{table*}[t]
\centering
\scriptsize
\setlength{\tabcolsep}{3pt}
\caption{\textbf{Ablation as paired differences to the reference} (variant minus \method on the same clips) with 95\% bootstrap intervals over clips. Rank-1 differences are computed per query clip, each within its own variant's gallery.}
\label{tab:s_ablation_delta}
\begin{tabular}{lcccccc}
\toprule
Variant & OSNet R1 & TransReID R1 & CSCI R1 & SPS$\times$100 & $\Delta E_\mathrm{warp}{\times}10^3$ & $\mathcal{A}_\mathrm{cons}$ \\
\midrule
w/o LMM filter & -4.75 {\tiny[-13.50, +3.88]} & -6.54 {\tiny[-13.20, -0.74]} & -7.65 {\tiny[-16.17, +0.43]} & -0.82 {\tiny[-1.15, -0.53]} & -0.00 {\tiny[-0.19, +0.19]} & -2.89 {\tiny[-4.66, -1.28]} \\
w/o temporal completion & +1.99 {\tiny[-1.81, +5.77]} & +2.67 {\tiny[-1.11, +6.69]} & +3.98 {\tiny[-2.63, +10.71]} & +0.02 {\tiny[-0.00, +0.04]} & -0.00 {\tiny[-0.07, +0.07]} & +0.82 {\tiny[+0.24, +1.44]} \\
neutral identity prompt & +0.41 {\tiny[-4.46, +4.75]} & +2.11 {\tiny[-1.55, +6.23]} & +4.57 {\tiny[-3.50, +12.70]} & +0.04 {\tiny[+0.01, +0.07]} & -0.51 {\tiny[-0.70, -0.32]} & +1.62 {\tiny[-0.30, +3.21]} \\
factual identity prompt & +1.36 {\tiny[-3.51, +6.06]} & +5.71 {\tiny[+1.74, +9.98]} & +8.00 {\tiny[+0.78, +15.05]} & +0.00 {\tiny[-0.04, +0.04]} & -0.14 {\tiny[-0.36, +0.10]} & +2.10 {\tiny[+0.55, +3.75]} \\
raw video as context & +11.09 {\tiny[+2.64, +19.33]} & +13.28 {\tiny[+6.37, +21.04]} & +11.92 {\tiny[+2.47, +21.45]} & +0.55 {\tiny[+0.49, +0.61]} & -0.50 {\tiny[-0.72, -0.29]} & +6.50 {\tiny[+4.97, +8.21]} \\
context scale 0 & -10.53 {\tiny[-17.72, -4.32]} & -10.78 {\tiny[-17.86, -5.01]} & -14.41 {\tiny[-22.79, -6.96]} & -40.80 {\tiny[-42.24, -39.38]} & -6.59 {\tiny[-7.73, -5.53]} & -48.61 {\tiny[-51.92, -45.16]} \\
\bottomrule
\end{tabular}
\end{table*}

\Cref{tab:s_seeds} reports the sensitivity of \method to the generation seed on the five predeclared hard Belvedere clips.

\end{document}